\documentclass[lettersize,journal]{IEEEtran}
\usepackage{capt-of}
\usepackage{algorithmic}
\usepackage{array}
\usepackage[caption=false,font=normalsize,labelfont=sf,textfont=sf]{subfig}
\usepackage{textcomp}
\usepackage{stfloats}
\usepackage{url}
\usepackage{verbatim}
\usepackage{graphicx}
\usepackage{amssymb}
\usepackage{amsmath}
\usepackage{xcolor} 

\definecolor{mycustomblue}{rgb}{0.21,0.49,0.74} 
\definecolor{mycustomblue1}{rgb}{0.21,0.49,0.74} 
\usepackage{hyperref}
\hypersetup{
  colorlinks=true,      
  linkcolor=mycustomblue1,       
  citecolor=mycustomblue1,      
  urlcolor=mycustomblue,        
  pdfborder={0 0 0}     
}
\usepackage{multirow}
\usepackage{booktabs}
\usepackage{pifont}
\def\BibTeX{{\rm B\kern-.05em{\sc i\kern-.025em b}\kern-.08em
    T\kern-.1667em\lower.7ex\hbox{E}\kern-.125emX}}
\usepackage{balance}

\begin{document}

\title{Distilling Multi-view Diffusion Models into 3D Generators}

\author{Hao Qin, Luyuan Chen, Ming Kong\textsuperscript{*}, Mengxu Lu, Qiang Zhu
	\thanks{\textsuperscript{*} Corresponding author.}
	\thanks{Hao Qin, Ming Kong, Mengxu Lu, and Qiang Zhu are with School of Computer Science and Technology, Zhejiang University,  Hangzhou 310027, China (e-mail: qinbaigao@gmail.com; zjukongming@zju.edu.cn; lumengxu@zju.edu.cn; zhuq@zju.edu.cn).}
	\thanks{Luyuan Chen is with School of Computer, Beijing Information Science and Technology University, Beijing 100005, China (email: chenly@bistu.edu.cn).}}

\markboth{Journal of \LaTeX\ Class Files,~Vol.~14, No.~8, August~2021}%
{Shell \MakeLowercase{\textit{et al.}}: A Sample Article Using IEEEtran.cls for IEEE Journals}

\twocolumn[{%
            \renewcommand\twocolumn[1][]{#1}
            \maketitle
            \begin{center}
                \centering
            \vspace{-10pt}
                \includegraphics[width=0.99\textwidth]{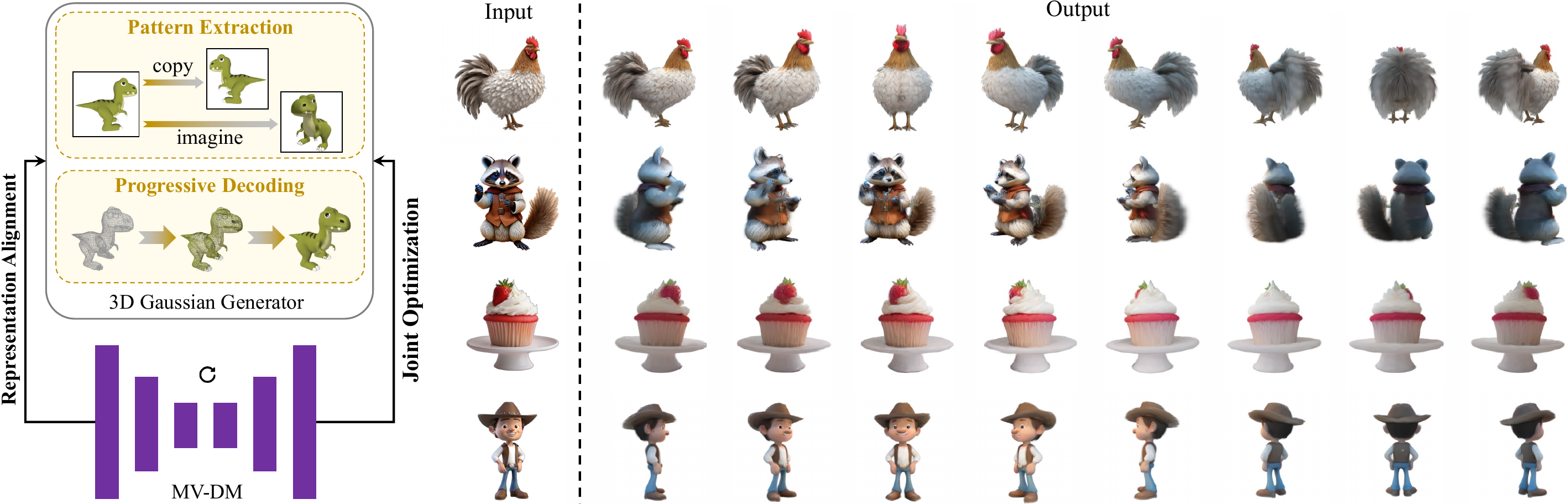}
                \captionof{figure}{DD3G can distill visual knowledge from MV-DM into the 3D Gaussian generator to achieve rapid and generalized high-quality 3D generation.}
                \vspace{5pt}
                \label{fig1}
            \end{center}
        }]

\begingroup
\renewcommand{\thefootnote}{}
\footnotetext{\textsuperscript{*} Corresponding author.}
\footnotetext{Hao Qin, Ming Kong, Mengxu Lu, and Qiang Zhu are with School of Computer Science and Technology, Zhejiang University,  Hangzhou 310027, China (e-mail: haoqin@zju.edu.cn; zjukongming@zju.edu.cn; lumengxu@zju.edu.cn; zhuq@zju.edu.cn).}
\footnotetext{Luyuan Chen is with School of Computer, Beijing Information Science and Technology University, Beijing 100005, China (email: chenly@bistu.edu.cn).}
\endgroup

\begin{abstract}
We introduce DD3G, a formulation that Distills a multi-view Diffusion model (MV-DM) into a 3D Generator using gaussian splatting. DD3G compresses and integrates extensive visual and spatial geometric knowledge from the MV-DM by simulating its ordinary differential equation (ODE) trajectory, ensuring the distilled generator generalizes better than those trained solely on 3D data. Unlike previous amortized optimization approaches, we align the MV-DM and 3D generator representation spaces to transfer the teacher’s probabilistic flow to the student, thus avoiding inconsistencies in optimization objectives caused by probabilistic sampling. The introduction of probabilistic flow and the coupling of various attributes in 3D Gaussians introduce challenges in the generation process. To tackle this, we propose PEPD, a generator consisting of Pattern Extraction and Progressive Decoding phases, which enables efficient fusion of probabilistic flow and converts a single image into 3D Gaussians within 0.06 seconds. Furthermore, to reduce knowledge loss and overcome sparse-view supervision, we design a joint optimization objective that ensures the quality of generated samples through explicit supervision and implicit verification. Leveraging existing 2D generation models, we compile 120k high-quality RGBA images for distillation. Experiments on synthetic and public datasets demonstrate the effectiveness of our method. Our project is available at: \url{https://qinbaigao.github.io/DD3G_project/}. 

\end{abstract}


\IEEEpubidadjcol

\section{Introduction}
\IEEEPARstart{W}{ith} the rapid development of 2D-AIGC \cite{rombach2022high} and 3D Gaussian Splatting \cite{kerbl20233d} technologies, there is a significant opportunity for the automated generation of 3D assets from a single image. However, a key challenge that has persisted is the scarcity of high-quality 3D data. 3D generators trained exclusively on 3D data \cite{huang2025learning}, although capable of rapid generation, typically exhibit limited generalization capabilities and lack a proper understanding of plausible 3D shapes. 

Thanks to the rich visual knowledge in pre-trained diffusion models \cite{wang2023imagedream, voleti2025sv3d}, the idea of constructing 3D objects using Score Distillation Sampling (SDS) \cite{poole2022dreamfusion} has recently gained widespread attention \cite{tang2023dreamgaussian, yi2024gaussiandreamer}. Despite significant progress, this approach is still limited by the complex, sample-wise optimization process, which requires substantial computational resources and time during the generation process. To improve generation speed, several studies adopt a two-stage 3D generation strategy: first producing multi-view images using MV-DM, then employing them for 3D reconstruction or generation \cite{li2024instant3d, tang2025lgm}. This approach significantly accelerates the generation process, but it is prone to error accumulation and often leads to voids and subtle shadow artifacts in the generated samples. In addition, repeated inference of the denoiser within MV-DM during the generation process undermines the real-time performance of this strategy. 

\IEEEpubidadjcol
Recently, some researchers have attempted to optimize the triplane generator directly using pre-trained visual models through amortized optimization \cite{amos2022tutorial, li2024instant3d, xie2024latte3d}. This approach allows for the generation of triplets with a single forward pass during inference, ensuring high generation efficiency while avoiding the limitations of scarce training data. Essentially, this method transfers the visual knowledge from the pre-trained model to the triplane generator, providing a reference for the automated generation of 3D Gaussians. However, this high-dimensional and complex knowledge mapping is undoubtedly a challenging task. Additionally, existing amortized optimization methods often lose latent probabilistic information during training, preventing complete alignment between the representation spaces of teacher and student models and thus hindering the knowledge distillation process.

To address the aforementioned challenges, we propose DD3G, which achieves rapid and generalized \textit{single image-to-3D Gaussians} generation by distilling an MV-DM into a 3D generator. During inference, the MV-DM is no longer required; instead, we rely solely on the distilled knowledge embedded within the 3D Gaussian generator to achieve feed-forward 3D generation. By doing so, we eliminate error accumulation and improve real-time performance compared to the conventional two-stage generation strategy, while the extensive visual knowledge retained in the MV-DM ensures the generalizability of the distilled model. Specifically, by simulating the ODE trajectory using the DDIM sampler \cite{song2020denoising}, we perform inference on the MV-DM to obtain \{\textit{N, C, II, OI}\} quadruples, where \textit{N}, \textit{C}, \textit{II}, and \textit{OI} represent, respectively, the initial noise sampled from a Gaussian distribution, the camera poses, the input image, and the output multi-view images. These quadruples are then employed in a weakly-supervised training regime for the 3D generator. The deterministic nature of the DDIM sampler ensures a one-to-one correspondence between \textit{N-C-II} and \textit{OI}, thereby aligning the representation space of the MV-DM and the 3D generator. Moreover, incorporating \textit{N} and \textit{C} allows the 3D generator to learn the probabilistic flow within the MV-DM, providing the essential condition for addressing the ill-posed nature of 3D generation.

The weakly-supervised distillation scenario introduces novel challenges in designing generators and optimization objectives. To better facilitate the conversion from \textit{N-C-II} to 3D Gaussians, we propose a novel 3D Gaussian generator named PEPD, which formulates the generation of 3D Gaussians as a 2D-to-3D lifting task. As illustrated in Fig.~\ref{fig1}, PEPD consists of two phases: Pattern Extraction (PE) and Progressive Decoding (PD). In the PE phase, to efficiently capture probabilistic flows in MV-DM and provide generalized guidance for subsequent generation, we map \textit{N} and \textit{C} under the guidance of \textit{II} to obtain the fundamental patterns for lifting the image. In the PD phase, we decode features at multiple levels of a single-branch network via distinct mapping heads. This allows us to effectively capture diverse attributes, maintaining computational efficiency and mitigating the issues caused by coupling among unstructured attributes in 3D Gaussians. Moreover, the transformer-based architecture enhances the scalability of PEPD, thereby providing significant potential for DD3G.

Simple data distillation can lead to overly sparse supervisory signals and significant knowledge loss. To address this limitation, we design a joint optimization objective that incorporates explicit supervision and implicit verification through a curriculum learning strategy. Explicit supervision ensures the alignment of representations between the teacher and student models during distillation by simulating the ODE solver, enabling the student model to learn an effective probabilistic flow. However, simple explicit distance metrics are insufficient to fully map the complex nonlinear distributions fitted through multiple iterations in MV-DM into the student model \cite{lin2024sdxl}. Meanwhile, the sparse multi-view images \textit{OI} lack full supervision for 3D objects, leading to confusing visual artifacts for certain unseen views. In the field of diffusion acceleration, adversarial loss has been proven effective in mitigating knowledge loss when distilling complex distributions \cite{kang2024distilling, yin2024one}. Yet, the scarcity of 3D data makes it challenging to construct a high-quality 3D discriminator or a dense multi-view consistency verifier. Alternatively, we revisit the SDS loss, which fails to efficiently capture the probabilistic information in MV-DM. However, when used judiciously, it can serve as an excellent multi-view consistency verifier with effects similar to adversarial loss. Therefore, we adopt it as an implicit verification objective.

The collection of quadruples and the computation of the implicit verification objective require RGBA images. To this end, we collect 120k high-quality RGBA images, including those generated using 2D-AIGC techniques \cite{achiam2023gpt,esser2024scaling,kirillov2023segment} and front-facing renderings of objects from Objaverse \cite{deitke2023objaverse}. All images have been manually filtered to exclude low-quality data. With the rapid advancement of image-to-3D generation technologies, there is a concomitant increase in the demand for high-quality RGBA image data featuring individual objects. In response, we will release our dataset publicly to serve as a foundational resource for subsequent research. 

Our main contributions are as follows:

\begin{itemize}

    \item We propose a formulation, DD3G, which enables the distillation of an MV-DM into a 3D Gaussian generator, offering a novel and general solution for image-to-3D generation.

    \item We introduce a novel 3D Gaussian generator, PEPD, which incorporates both the Pattern Extraction and Progressive Decoding phases. PEPD efficiently learns the probabilistic flow within MV-DM and enables fast and generalized image-to-3D lifting.

    \item We develop a joint optimization objective by combining explicit supervision and implicit verification through curriculum learning, addressing the challenges of sparse supervisory signals and knowledge loss inherent in the data distillation process.

    \item We collect 120k high-quality RGBA image data, which provides a foundation for future research. Extensive experiments on synthetic and public datasets validate the effectiveness of our approach.

\end{itemize}

\section{Related Work}
\label{sec:related}
\begin{figure*}[t]
  \centering
   \includegraphics[width=\linewidth]{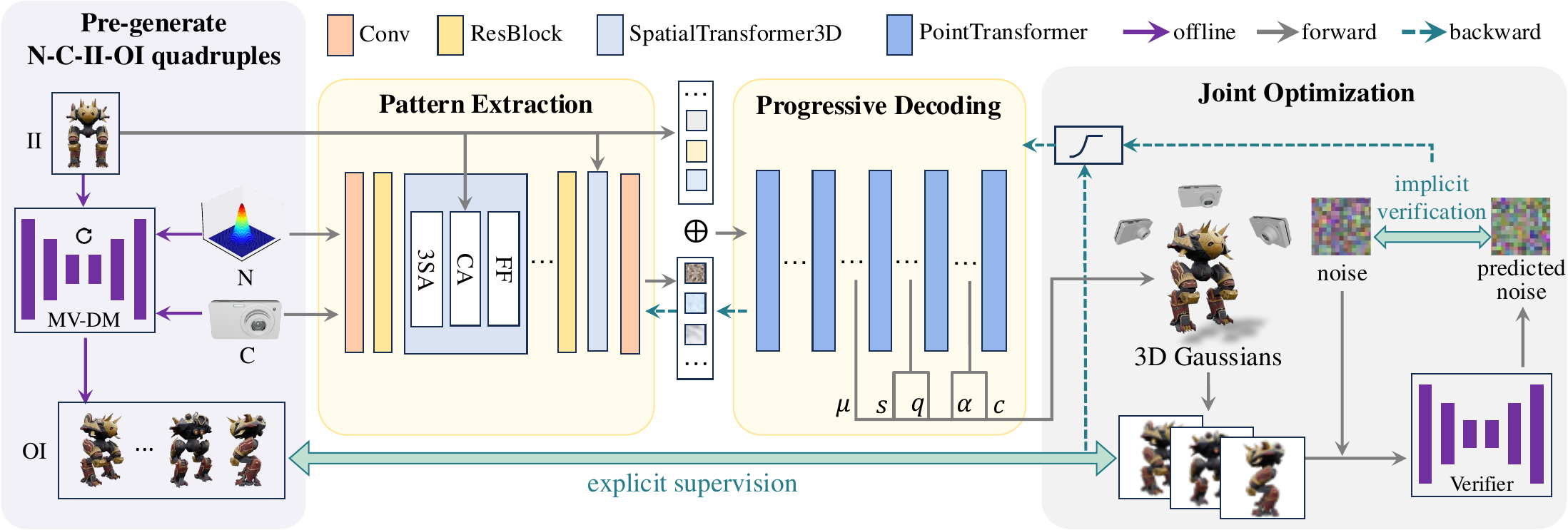}
   \caption{DD3G trains the 3D Gaussian generator PEPD to lift a single image into a 3D object. Given the offline collected \{\textit{N, C, II, OI}\} (noise, camera pose, input image, output multi-view images) quadruples as training samples, the Pattern Extraction (PE) phase extracts the lifting pattern of \textit{II} from random information \textit{NC}, as the general guidance for the Progressive Decoding (PD) phase to decouple the 3D Gaussian attributes progressively. Furthermore, the joint optimization objective that combines explicit supervision with implicit verification is formed to improve the quality of generated samples.}
   \label{fig2}
\end{figure*}

\subsection{3D Generation through Score Distillation}
\label{sec:2.1}

SDS \cite{poole2022dreamfusion}, also known as Score Jacobian Chaining (SJC) \cite{wang2023score}, is an optimization method that distills knowledge from pre-trained diffusion models. Given the rich visual knowledge embedded in these pre-trained models, SDS enables generalized 3D generation. To further improve the quality and efficiency of the generation process, researchers have proposed various score distillation methods following enhancements to SDS \cite{ma2025scaledreamer, wang2024prolificdreamer, zhuo2024vividdreamer}. These methods largely adhere to the same optimization paradigm:
\begin{equation}
    \mathcal{L}_{SD} = \mathbb{E}_{t, \epsilon, \tilde{c}} \left[ w(t) \left( \epsilon_{\phi}(x_t; t, y, \tilde{c}) - \hat{\epsilon} \right) \right],
    \label{equ1}
\end{equation}
where, $\hat{\epsilon}$ can be the noise predicted by the original or fine-tuned U-Net. Different design choices lead to varying generation outcomes. Despite significant progress, this sample-wise optimization approach still requires substantial computational resources.

\subsection{Feed-forward 3D Generation}
\label{sec:2.2}

To achieve fast and automated 3D generation, some studies have attempted to train a 3D generator. Initially, researchers learned priors for 3D objects of fixed categories through an encoder-decoder structure \cite{xu2019disn, mescheder2019occupancy}. As the architectures of generative models evolved, some works realized multi-type 3D generation by designing multi-parameter models \cite{xu2023dmv3d, kwon2025text2avatar, yin2025shapegpt, hong2023lrm}. However, these approaches have been limited by the scarcity of 3D data. In addition, to mitigate the time constraints caused by per-sample optimization in score distillation, works such as \cite{amos2022tutorial, li2024instant3d, xie2024latte3d} have converted the optimization objective of score distillation from a single 3D object to a 3D generator. These methods achieve prompt-amortized feed-forward 3D generation but suffer from the loss of probabilistic flow in pre-trained visual models. VFusion3D \cite{han2025vfusion3d}, which uses samples generated by video diffusion models \cite{girdhar2023emu} to train LRM \cite{hong2023lrm}, faces a similar issue. Our method also falls under feed-forward 3D generation, and it achieves fast 3D generation without the loss of probabilistic flow.

\subsection{Generative Gaussians Splatting}
\label{sec:2.3}

Thanks to its excellent rendering speed and representation quality, 3D Gaussian Splatting (3DGS) \cite{kerbl20233d} has recently attracted significant attention and made substantial progress \cite{chen2024survey, yu2024mip}. The original 3DGS requires the reconstruction of 3D Gaussians from multiple images. Later studies begin exploring methods to obtain high-quality 3D Gaussians using only a single image. TGS \cite{zou2024triplane}, AGG\cite{xu2024agg}, and BrightDreamer \cite{jiang2024brightdreamer} indirectly obtain 3D Gaussians by combining point clouds with triplanes. SI \cite{szymanowicz2024splatter} maps the input image to one 3D Gaussian per pixel using 2D operators. GVGEN \cite{he2025gvgen} and GaussianCube \cite{zhang2024gaussiancube} apply structural constraints on the 3D Gaussians to align them with existing 3D generative models. \textit{Single image-to-3D Gaussians} is an ill-posed task and requires external knowledge, with existing approaches relying on limited 3D data. Considering that pre-trained MV-DMs contain rich 2D and 3D visual knowledge, we attempt to leverage this knowledge to achieve high-quality 3D Gaussian generation. To the best of our knowledge, we are the first to attempt distilling knowledge from MV-DM into a 3D Gaussian generator.

\section{Method}
\label{sec:method}

Our goal is to transfer visual knowledge from the pre-trained MV-DM $U_{base}$ to the 3D generator PEPD $G_\theta$ via knowledge distillation. This process is designed to model the distribution of image-to-3D lifting and achieve rapid, high-quality 3D Gaussian generation. We begin with a brief overview of the preliminary concepts. Next, we provide a detailed explanation of the two phases and the underlying mechanisms of $G_\theta$. We then examine explicit supervision and implicit verification, introducing the joint optimization objective through a curriculum learning strategy. Finally, we outline the data collection and inference processes. 

\subsection{Preliminaries}
\label{sec:3.1}

\textbf{Denoising Diffusion Implicit Models (DDIM)} \cite{song2020denoising} build upon the foundation laid by Denoising Diffusion Probabilistic Models (DDPM) \cite{ho2020denoising} but introduce significant modifications to the sampling process. It reinterprets the reverse diffusion process as solving an ODE, enabling deterministic computation of the reverse trajectory. This reinterpretation not only removes the inherent randomness of DDPM but also streamlines generation, yielding a more efficient sample reconstruction. The deterministic sampling process provides prerequisites for $G_\theta$ to learn the probability flow in $U_{base}$.

\textbf{3D Gaussian Splatting (3DGS)} \cite{kerbl20233d} represents a 3D scene with a set of anisotropic 3D Gaussians $\mathcal{G}$. Each 3D Gaussian $\mathcal{G}_{i}$ is composed of its position $\mu_i \in \mathbb{R}^3$, scaling factor $s_i \in \mathbb{R}^3$, rotation quaternion $q_i \in \mathbb{R}^4$, color information $c_i\in \mathbb{R}^c$, and opacity $\alpha_i \in \mathbb{R}$, i.e., $\mathcal{G}_{i}=\left\{\mu_{i}, s_{i},c_{i}, \alpha_{i}, q_{i}\right\}$. The tile-based CUDA rasterizer allows real-time differentiable rendering of 3DGS \cite{jiang2024gs}. However, the various unstructured attributes of 3D Gaussians are coupled, making it challenging to generate all 3D Gaussians at once.

\textbf{Score Distillation Sampling (SDS)} \cite{poole2022dreamfusion} bridges the gap between pre-trained 2D diffusion models and 3D generation, with its core idea illustrated in Eq.~\ref{equ1}. The effectiveness of SDS is heavily influenced by the noise timestep \textit{t}; specifically, a higher noise level enables the 3D model to capture coarse-grained features, whereas a lower noise level allows it to focus on fine-grained details \cite{yi2024diffusion}.

\subsection{PEPD}
\label{sec:3.2}

As shown in Fig.~\ref{fig2}, after offline collecting a sufficient number of \{\textit{N, C, II, OI}\} quadruples using a deterministic ODE solver, we employ the two-phase $G_\theta$ to facilitate the transformation from a single image to 3D Gaussians $\mathcal{G}$:
\begin{equation}
    \mathcal{G} = p_2(II \oplus p_1(II, N, C)),
\end{equation}
where $\oplus$ denotes concatenation, and $p_1$ and $p_2$ correspond to the first and second phases of $G_\theta$, respectively.

\textbf{Pattern Extraction (PE):} Unlike original diffusion models~\cite{rombach2022high}, the reverse trajectory of MV-DM is influenced by the camera parameters. Therefore, we first employ the Plücker ray embedding \cite{xu2023dmv3d} to densely encode \textit{C} in the quadruple and concatenate it with \textit{N} to form a random variable \textit{NC}. Let the number of sampled noise instances be $n$, so that \textit{NC} contains $n$ sub-random variables, each corresponding to a viewpoint. In order for $G_\theta$ to learn the probabilistic flow from $U_{base}$, we need to integrate \textit{NC} into the generative process by some sensible means. 

During the image-to-3D lifting process, the unseen regions must be inferred from the visible regions in the input image. Considering the varying utilization patterns of the visible regions by different types of objects, such as symmetric objects that can directly replicate visible regions, while irregular objects cannot overly rely on them, we design the Pattern Extraction phase to perform an initial mapping of \textit{NC} under the guidance of \textit{II}, providing direction for the subsequent generation process. Specifically, we employ the Cross Attention (CA) layer to inject image information into \textit{NC}, and utilize the 3D Self-Attention (3SA) layer \cite{shi2023mvdream} to enable interaction between the $n$ sub-random variables in \textit{NC}, thereby ensuring the model’s capacity to map space features. At both the input and output sides of the Pattern Extraction, we apply convolutional layers to transform feature dimensions; in the intermediate part, we stack ResBlocks and SpatialTransformer3D layers, which consist of 3SA, CA, and the Feed-Forward (FF) layer, to extract the lifting pattern. The pattern tokens are concatenated with image tokens and serve as the input for the Progressive Decoding phase.

\textbf{Progressive Decoding (PD):} The various attributes of 3D Gaussians are interrelated, and different combinations of these attributes can produce the same rendered image. When using 2D images for supervision, multiple reasonable gradient descent directions can hinder the expected optimization of these attributes. To decouple these unstructured attributes, previous work utilized a dual-branch structure to transform the prediction of positional information into a point cloud generation task, thereby separating $\mu_i$ from the other attributes \cite{zou2024triplane, xu2024agg, jiang2024brightdreamer}. However, this dual-branch structure not only reduces the model's computational efficiency but also prevents any interaction between $\mu_i$ and the other attributes.

\begin{figure}[t]
  \centering
   \includegraphics[width=0.98\linewidth]{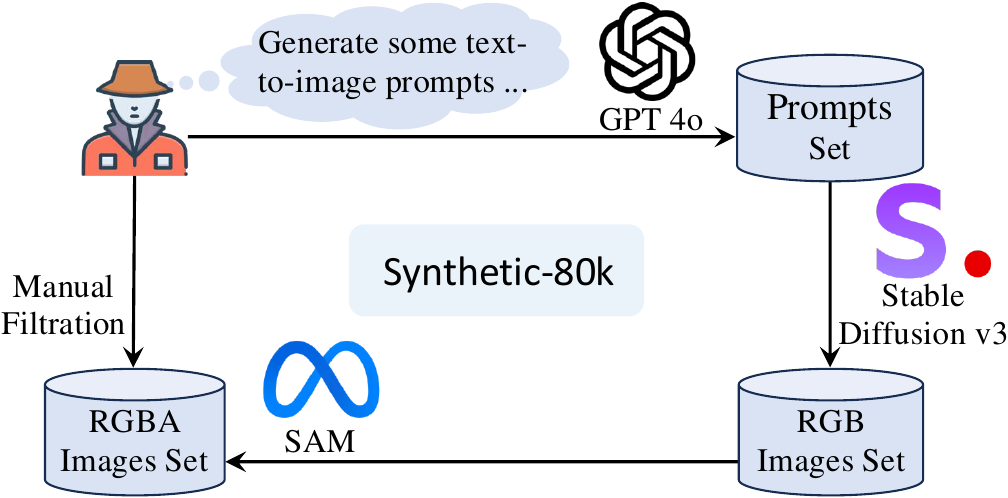}
   \caption{Overview of the synthetic image collection process. We adopt the same method as in \cite{zou2024triplane} to extract the foreground objects in images.}
   \label{fig3}
\end{figure}

\begin{figure*}[t]
  \centering
   \includegraphics[width=\linewidth]{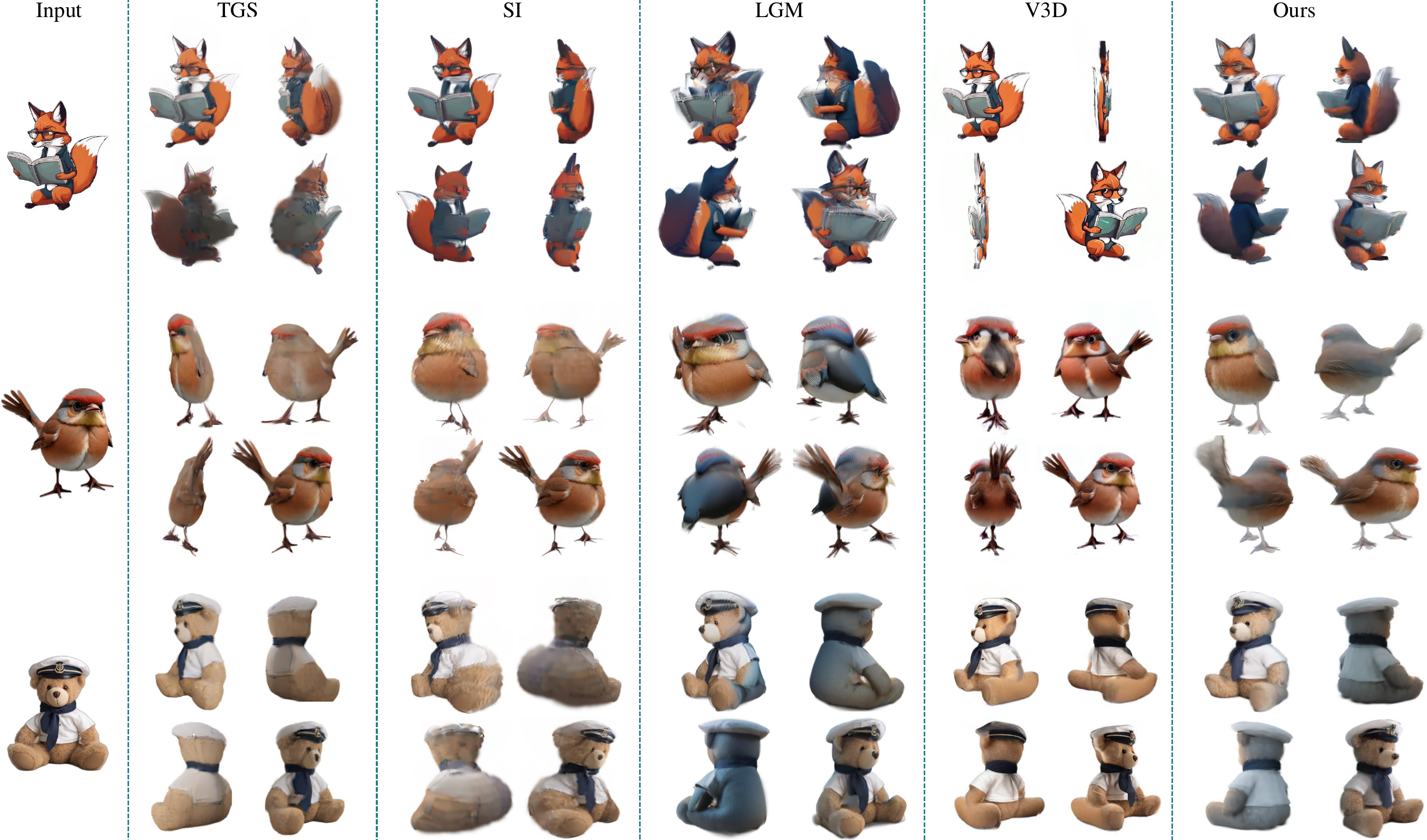}
   \caption{Qualitative comparisons between DD3G and other baselines. Our method achieves significant advantages in overall geometric consistency, primarily due to the efficient utilization of visual knowledge from MV-DM.}
   \vspace{-5pt}
   \label{fig4}
\end{figure*}

To address this, we design a single-branch network that decouples different attributes through a progressive decoding strategy. Specifically, we construct the backbone of the decoding process by stacking PointTransformer \cite{wang2019dgl} blocks, and then decode features from different levels of the network using several simple MLP decoders to obtain various attributes. Considering the meaning of each attribute, we perform progressive decoding in a manner similar to how humans construct 3D objects. As shown in Fig.~\ref{fig2}, we first decode $\mu_i$ to obtain the basic shape of the 3D object, then decode $s_i$ and $q_i$ respectively to capture the object's outline, and finally decode $c_i$ and $\alpha_i$ to retrieve detailed information. The combination of progressive decoding and a single-branch structure ensures that different attributes have relatively independent optimization paths while enabling sufficient interaction between them.

Multiple 3D Gaussians in an object may have similar attributes, so we choose to decode multiple 3D Gaussians from a single output token to improve the detail of the generated samples. We do not apply any structural constraints to the 3D Gaussians to avoid reducing their representational capacity. Experiments in Fig.~\ref{fig6} show that PEPD is more effective in the distillation scenario compared to the baseline. The specific structure of PEPD is listed in Supplementary Materials.

\subsection{Joint Optimization}
\label{sec:3.3}

We train $G_\theta$ by minimizing two optimization objectives: explicit supervision and implicit verification.

\textbf{Explicit Supervision:} The explicit supervision objective aims to encourage $G_\theta$ to match the large scale structure of the output of $U_{base}$ on a fixed dataset of many \{\textit{N, C, II, OI}\} quadruples. Essentially, it achieves knowledge distillation by simulating the ODE trajectory of $U_{base}$:
\begin{equation}
    \mathcal{L}_{\text{distill}}^{\text{ODE}} = \mathbb{E}_{N, C, II} \left[ d\left( \mathcal{R}\left( G_{\theta}(N, C, II), C \right), OI \right) \right],
\end{equation}
where, $\mathcal{R}$ represents differentiable rasterization rendering, $d$ represents the distance metric function, and in DD3G we use Mean Squared Error (MSE) and Learned Perceptual Image Patch Similarity (LPIPS) \cite{zhang2018unreasonable}, so the overall explicit supervision objective can be expressed as:
\begin{equation}
    \mathcal{L}_{\text{ES}} = \lambda_1 \mathcal{L}_{\text{MSE}}^{\text{ODE}} + \lambda_2 \mathcal{L}_{\text{LPIPS}}^{\text{ODE}},
\end{equation}
where, $\lambda$ is the weight coefficient. Although explicit supervision can achieve knowledge distillation on its own, we observe that it causes geometric inconsistencies in the generated 3D objects, which results in lower multi-view coherence. In the field of diffusion acceleration, Lin et al. \cite{lin2024sdxl} find that transferring the complex distribution from the teacher model to the student model using distance metrics such as MSE, L1, and LPIPS leads to knowledge loss. They also point out that adversarial objectives can effectively mitigate this issue. Based on this, we aim to incorporate adversarial objectives into 3D distillation. However, considering the limited data and the complexity of unstructured representations, training a robust 3D discriminator presents significant challenges.

\textbf{Implicit Verification:} SDS assesses the validity of an image by predicting the noise added on its feature map. We observe that, when the noise intensity is controlled at a low level, SDS can implicitly verify the sample within the feature space, ensuring it aligns with the true distribution without imposing a specific form. This is conceptually similar to the discriminator loss. Therefore, we employ MV-DM as a verifier and use multi-view SDS to construct an implicit verification objective, addressing challenges such as knowledge loss introduced by explicit supervision and the poor reasoning in invisible regions caused by sparse supervisory views in \textit{OI}:
\begin{align}
    & \mathcal{L}_{\text{IV}} = \mathbb{E}_{t, II, \epsilon, \tilde{c}} \left[ w(t) \left( \epsilon_{\phi} (z_t; t, II, \tilde{c}) - \epsilon \right) \right], \nonumber \\
    & \text{s.t.} \quad z_t = forward(\mathcal{R}(G_{\theta}(N, C, II), \tilde{c}), t),
\end{align}
where, $w(t)$ is a weighting function, $\tilde{c}$ represents the random camera poses, and $forward$ denotes the encoding and forward process of MV-DM. To ensure that the implicit verification objective does not disrupt the coherent information within the generated samples, we set the noise-adding timestep $t$ within a relatively small range.

\textbf{Curriculum Learning:} In the early stages of model training, the quality of generated samples is poor. This makes it difficult for the implicit verification—with fewer noise-adding timesteps—to provide effective gradients, whereas the more fundamental explicit supervision objective remains unaffected. To address this, we gradually introduce implicit verification during the training process while keeping explicit supervision unchanged:
\begin{align}
    & \mathcal{L}_{\text{DD3G}} = \mathcal{L}_{\text{ES}} + \beta_i \mathcal{L}_{\text{IV}}  + \mathcal{L}_{\text{on}}, \nonumber \\
    \text{s.t.} & \quad \beta_i = \min \left( 0.5 \ast \left( e^{{i} \slash {I}} - 1 \right), \eta \right),
\end{align}
where, $i$ and $I$ represent the current iteration number and the total number of iterations, respectively, $\eta$ is a hyperparameter used to control the upper limit of the weight. $\mathcal{L}_{\text{on}}$ is the opacity regularization term to prevent the generation of an excessive number of transparent 3D Gaussians:
\begin{align}
\mathcal{L}_{on} = -\frac{1}{M}\sum_{i=1}^{M}\log \alpha_i \quad \text{with } \alpha_i \in (0, 1) \quad \forall i,
\end{align}

\textbf{Random Background Color:} We find that if the background color is kept white during training, $G_{\theta}$ tends to generate many white 3D Gaussians around the target object. These white 3D Gaussians not only lead to computational waste but also cause the model parameters to fall into local optima. To address this issue, we adopt random background colors for the implicit verification objective:
\begin{equation}
b g c=255 *\left(1-(1-\zeta)^{3}\right),
	\label{eq5}
\end{equation}
where, $bgc$ is the the background color and $\zeta \sim U(0,1)$.

\subsection{Collecting RGBA Images}
\label{sec:3.4}

In the distillation process of DD3G, the collection of quadruples and the computation of the implicit verification objective both rely on RGBA images. However, the quality of existing RGBA datasets is generally poor and unsuitable for 3D generation. To address this, we collect 120k high-quality RGBA images, including 80k synthetic images and 40k images rendered from objects in the Objaverse~\cite{deitke2023objaverse} dataset. The detailed process of collecting synthetic images is illustrated in Fig.~\ref{fig3}. All images are manually verified to ensure data quality. The contents of these images cover a variety of everyday objects, plants, animals, as well as various virtual objects and cartoon characters. We will make all the data publicly available to promote future research in image-to-3D generation.

\subsection{Inference}
\label{sec:3.5}

After the distillation of our DD3G, $G_\theta$ is optimized to quickly generate high-quality 3D Gaussians from a single image. Users can simply randomly select a set of camera poses and concatenate them with Gaussian noise, then perform inference on PEPD with the input image to obtain the corresponding 3D object.

\begin{table*}[t]
  \centering
    \setlength{\tabcolsep}{3.5mm}{
    
     \caption{Quantitative Comparison Between DD3G and Baselines on Google Scanned Objects and RTMV-bricks. ``w/o MV-DM'' Indicates that MV-DM is not Required During the Generation Process.}
     \vspace{-3pt}
\label{tab1}
    
    \begin{tabular}{c|c|ccc|ccc|c}
    \toprule
    \multirow{2}{*}{Method} & \multirow{2}{*}{w/o MV-DM} & \multicolumn{3}{c|}{Google Scanned Objects} & \multicolumn{3}{c|}{RTMV-bricks} & \multirow{2}{*}{Generation Time (s) $\downarrow$} \\
    & & PSNR $\uparrow$ & SSIM $\uparrow$ & LPIPS $\downarrow$ & PSNR $\uparrow$ & SSIM $\uparrow$ & LPIPS $\downarrow$ & \\ \midrule
    SI \cite{szymanowicz2024splatter} & \ding{51} & 16.03 &  0.815&  0.201 & 11.76 & 0.582 & 0.319 & \textbf{0.03}  \\
    TGS \cite{zou2024triplane} & \ding{51} & 17.37 &  0.820&  0.189 & 11.85 & 0.591 & 0.310 & 0.11  \\ \midrule
    DG \cite{tang2023dreamgaussian} & \ding{55} & 13.77 & 0.801 & 0.254 & 10.22 &  0.569 &  0.332 & 106  \\
    LGM \cite{tang2025lgm} & \ding{55} & 15.60 &  0.823&  0.195& 11.83 &  0.584 &  0.297 & 3  \\ 
    V3D \cite{chen2024v3d} & \ding{55} & \underline{19.26} &  \underline{0.847} &  \underline{0.163} & \underline{14.05} &  \underline{0.659} &  \underline{0.212} & 174  \\ \midrule
    Ours & \ding{51} & \textbf{19.85} & \textbf{0.883} & \textbf{0.131} & \textbf{15.83} &  \textbf{0.702} & \textbf{0.168} & \underline{0.06}  \\
    \bottomrule
    \end{tabular}}
     \vspace{-5pt}

\end{table*}

\section{Experiments}
\label{sec:experiment}
\subsection{Experimental Setup}
\label{sec:4.1}

\textbf{Data:} We use 120k RGBA images for distillation and hold out 200 images as a test set. In addition, we also test on Google Scanned Objects (GSO) \cite{downs2022google} and RTMV-bricks \cite{tremblay2022rtmv}. To strike a balance between training cost and generation quality, we simulate the ODE of ImageDream \cite{wang2023imagedream} using 50 steps of DDIM, generating 2.8 million quadruples.

\textbf{Training:} During training, we set both $\lambda_1$ and $\lambda_2$ to 0.5 and $\eta$ to 0.4. $t$ is randomly selected from the range [20, 300], and the rotation angle of $\tilde{c}$ ranges from [-180, 180], while the pitch angle ranges from [-5, 5]. We use the Adam optimizer \cite{kingma2014adam} to update the parameters of PEPD with the learning rate set to 1e-5. The size of the rendered images is set to 256*256. To enhance the stability of training, we apply Exponential Moving Average (EMA) to the model parameters and clipped the gradients during backpropagation to the range (-5.0, 5.0). All scaling factors $s$ are adjusted to the range of (-9, -3) using the sigmoid function.  We train the model on a 6 A100-80G server, with the batch size set to 72. The model training and quadruple generation take approximately 2,600 GPU hours.

\begin{figure}[t]
  \centering
   \includegraphics[width=0.9\linewidth]{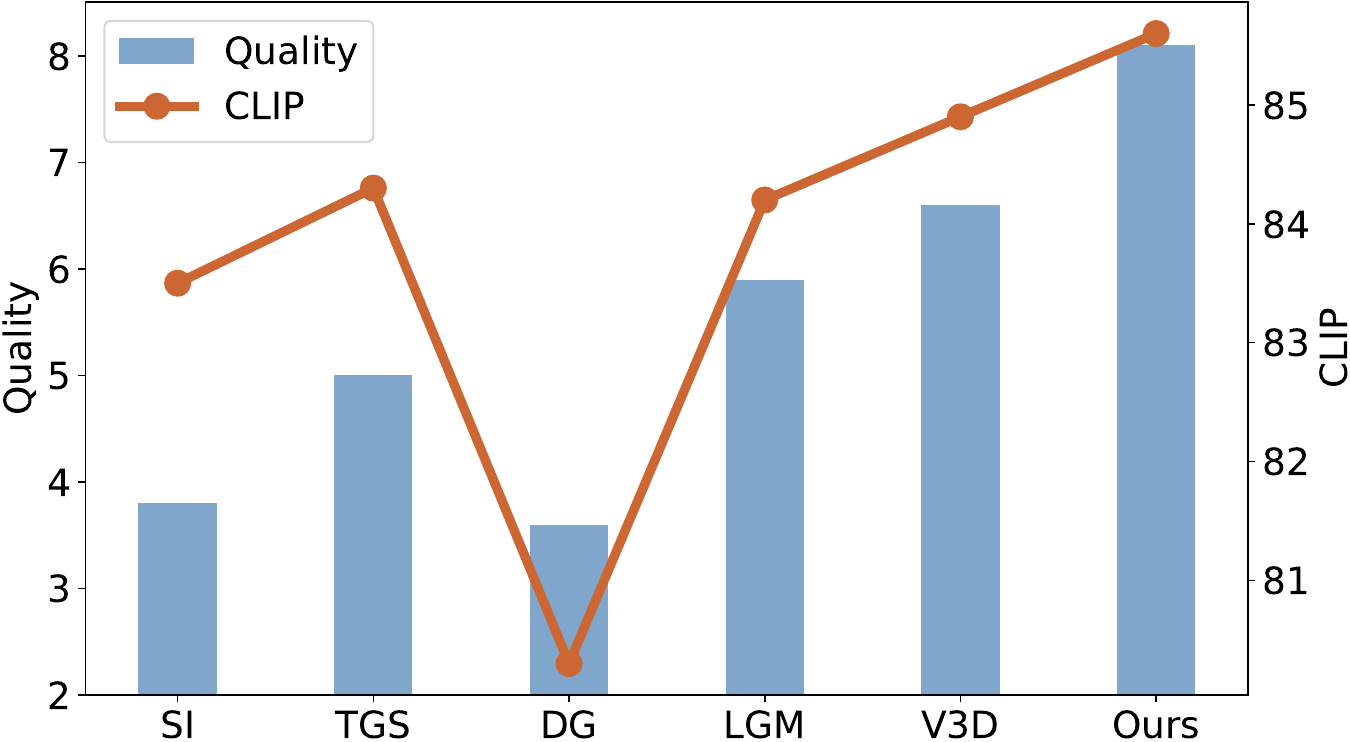}
   \caption{Illustration of CLIP Similarity and User Study. The samples generated by PEPD are more aligned with human preferences.}
   \label{fig5}
\end{figure}

\textbf{Progressive Decoding:} We weight and sum the features extracted from the outputs of adjacent PointTransformer blocks to form the input for a specific attribute's decoder. The weighting coefficients for the inputs of the $\mu$, $s$, $q$, $c$, and $\alpha$ decoders are respectively [0.15, 0.7, 0.15], [0.15, 0.7, 0.15], [0.15, 0.7, 0.15], [0.2, 0.8], and [0.2, 0.8]. Each token is decoded into 512 3D Gaussians, and the 3D object generated by PEPD contains a total of 157,184 3D Gaussians.

\textbf{Baselines and Metrics:} We compare our method with previous state-of-the-art \textit{single image-to-3D Gaussians} methods. Based on whether MV-DM is involved in the generation process, these methods can be categorized into two types: (1) SI \cite{szymanowicz2024splatter}, TGS \cite{zou2024triplane}; (2) DG \cite{tang2023dreamgaussian}, LGM \cite{tang2025lgm}, and V3D \cite{chen2024v3d}. In line with \cite{zou2024triplane, szymanowicz2024splatter}, we calculate Peak Signal-to-Noise Ratio (PSNR), Structural Similarity (SSIM), and perceptual loss (LPIPS) on GSO and RTMV-bricks. Since our work, as well as DG and LGM, assume that the input image's angle is not zero but a hidden variable, during testing, we manually rotate the generated samples to align their 0-degree view with the input image. Additionally, we follow \cite{tang2023dreamgaussian, tang2025lgm} to calculate the clip similarity \cite{radford2021learning} between the input image and the rendered images of the output sample's various views, and conduct a user study.

\subsection{Comparison with Baselines}
\label{sec:4.2}

\textbf{Qualitative Comparisons:} We compare the generated samples of PEPD trained via DD3G with those of the baselines, as shown in Fig.~\ref{fig4}. TGS and SI are feed-forward methods that offer fast generation speeds; however, the limited 3D training data restricts their ability to understand spatial structures, resulting in geometric distortions in the generated objects. LGM and V3D require the involvement of MV-DM during inference. While they produce better generation results from certain perspectives, they are prone to artifacts from other viewpoints. By directly distilling visual knowledge from MV-DM into the generator, we achieve efficient and geometrically consistent 3D generation.

\begin{figure}[t]
  \centering
   \includegraphics[width=0.96\linewidth]{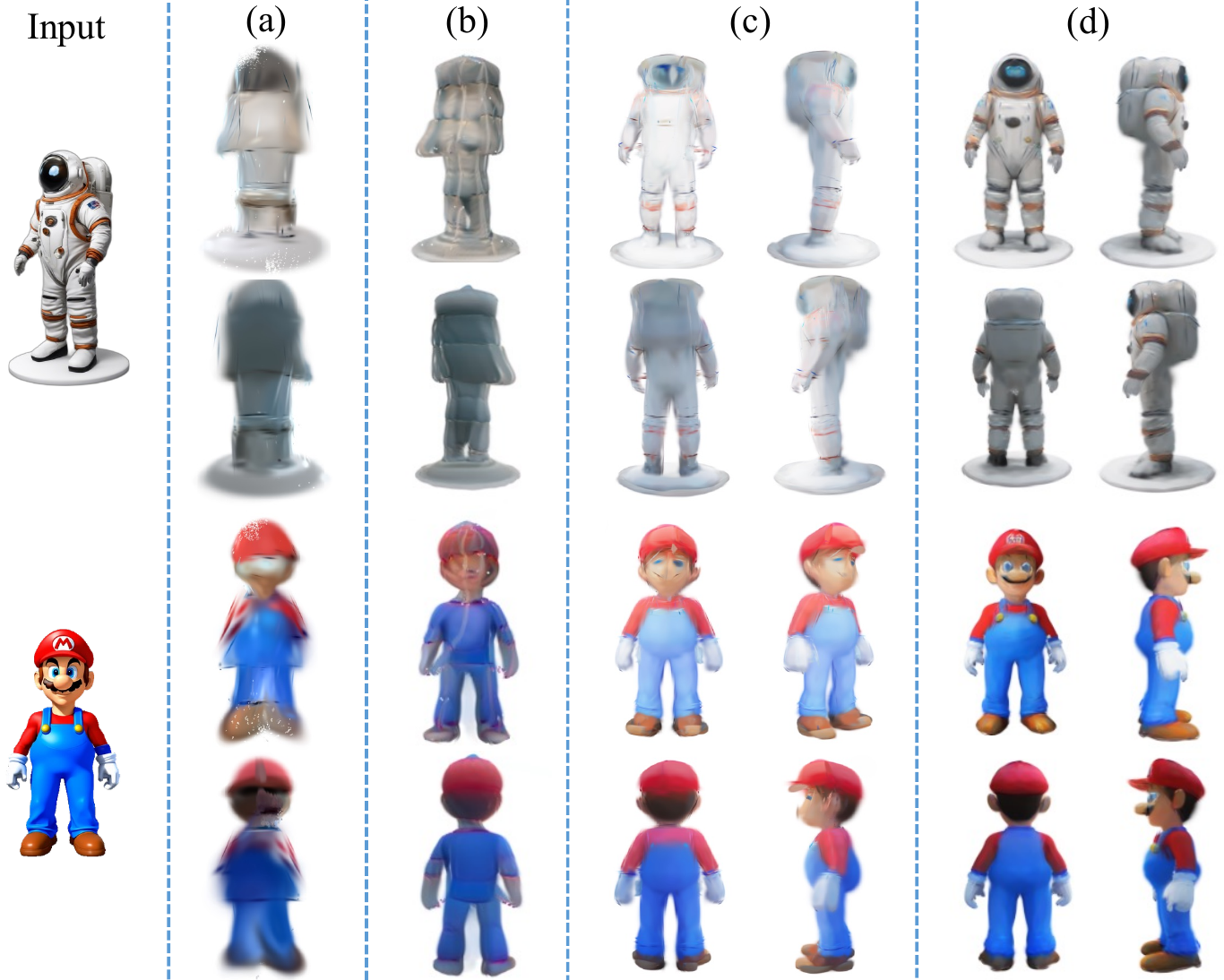}
   \caption{Ablation study of PEPD. We train different backbones in DD3G with the explicit supervision objective, which are (a) TGS, (b) SI, (c) PD, and (d) the full PEPD.}
   \vspace{-5pt}
   \label{fig6}
\end{figure}

\textbf{Quantitative Comparisons:} In addition, we conduct quantitative comparisons. First, we render 12 images from different viewpoints for each object in the test set and evaluate the quality of novel view synthesis in comparison with baselines, as shown in Tab.~\ref{tab1}. It can be observed that our method significantly outperforms all baselines across all metrics while maintaining fast generation speeds. Next, we calculate the CLIP score between the input image and the rendered images of the generated 3D objects using 200 synthesized images and conduct a user study. We randomly present 10 sets of generated samples from different methods to 47 volunteers, who are asked to rate the 3D objects based on their intuitive impressions, with a score of 1 representing very poor quality and 10 representing perfection. Fig.~\ref{fig5} shows the average CLIP score and quality score for each method, and it can be seen that our method best aligns with human preferences. The Supplementary Materials provide comparisons with generators of other representations.

\begin{table}
\begin{center}
\setlength{\tabcolsep}{3.6mm}
\caption{Quantitative Results on GSO of the Ablation Study in Fig.~\ref{fig6}.}
\vspace{-3pt}
\label{tab2}
\begin{tabular}{cccc}
    \toprule
    Group & PSNR $\uparrow$ & SSIM $\uparrow$ & LPIPS $\downarrow$  \\
    \midrule
    (a) & 15.88 &  0.781&  0.204   \\
    (b) & 16.95 &  0.793&  0.193  \\
    (c) & 18.16  & 0.831  &  0.175  \\ \midrule
    (d) & \textbf{19.85} & \textbf{0.883} & \textbf{0.131}  \\
    \bottomrule
\end{tabular}
\end{center}
\vspace{-8pt}
\end{table}

\subsection{Effectiveness of Each Component}
\label{sec:4.3}

To achieve efficient knowledge distillation, we propose a novel 3D Gaussian generator, PEPD, and design a joint optimization objective specifically tailored for DD3G. Here, we validate their effectiveness through ablation experiments. Fig.~\ref{fig6} demonstrates the impact of different network architectures on the generated samples when distillation is performed using the explicit supervision objective. Tab.~\ref{tab2} presents the quantitative experimental results on GSO for the various experimental groups illustrated in Fig.~\ref{fig6}.

\textbf{Effectiveness of PD:} Initially, we intend to directly use existing 3D Gaussian generators, such as TGS and SI. However, during the implementation, we observe that these generators are designed for 3D data and require knowledge of the camera pose of the input image during training. They also align the output with standard images rendered from 3D data. In our experiments, we fix the camera pose of the input image with the expectation that the network can automatically disregard the influence of the camera pose. As shown in Fig.~\ref{fig6}a and \ref{fig6}b, existing generators do not converge well in DD3G. We hypothesize that this is because the distillation process of DD3G is essentially a form of weakly-supervised learning, which cannot provide precise camera poses and supervisory images, leading to the failure of the components in 3D-data-oriented generators. Fig.~\ref{fig6}c shows the experimental results when ignoring probabilistic information and using only the PD. It can be observed that using PD alone achieves better generative samples than other network architectures, primarily due to the simplified single-branch structure and the progressive decoding strategy.

\textbf{Effectiveness of PE:} As shown in Fig.~\ref{fig6}c, when the probabilistic information from the teacher model is ignored, the generated samples, while geometrically reasonable, lose many fine details. During training, we observe that directly concatenating the probabilistic information with the input image results in an unstable training process. To efficiently integrate the probabilistic flow into the generator, we design the Pattern Extraction (PE) phase. The experimental results of distilling the full PEPD are presented in Fig.~\ref{fig6}d. It can be observed that with the introduction of PE, even when only explicit supervision is used, the model's generation capability improves significantly, with the generated samples exhibiting richer texture details.

\begin{figure}[t]
  \centering
   \includegraphics[width=0.96\linewidth]{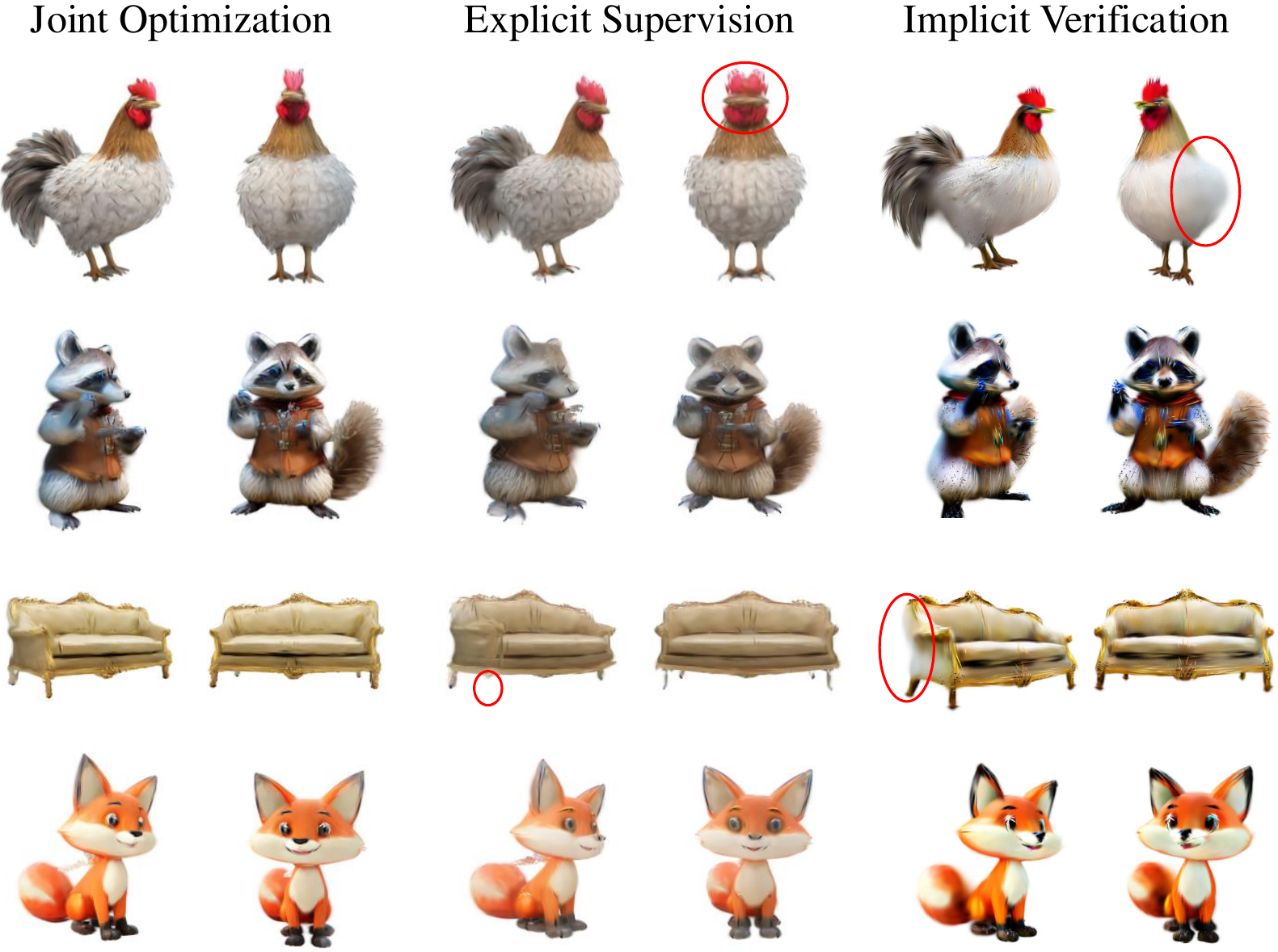}
   \caption{Ablation study of optimization objectives. The absence of implicit verification leads to unreasonable generated samples in certain views and weaker overall color. Meanwhile, the lack of explicit supervision results in the loss of details.}
   \label{fig7}
\end{figure}

\textbf{Effectiveness of Joint Optimization:} Furthermore, Fig.~\ref{fig7} and Tab.~\ref{tab3} present the experimental results obtained using the complete PEPD as the generator with different optimization objectives. It can be observed that, in the absence of the implicit verification objective, the model still ensures convergence. However, the generated samples exhibit confusing visual effects from certain viewpoints, and the overall color saturation is insufficient. When only implicit verification is used, the model fails to converge, likely due to the insufficient noise intensity, which fails to provide an effective optimization gradient for the generator. Therefore, we redesign the noise intensity according to the strategy in \cite{wang2024prolificdreamer} and conduct experiments, as shown in Fig.~\ref{fig7}. In the absence of explicit supervision, the generated samples tend to lose fine details, and regions that are not visible become blurred. We attribute this to the fact that, in addition to the oversaturation characteristics introduced by the SDS loss itself \cite{ma2025scaledreamer}, the lack of explicit supervision reduces the ability of the student model to align its representation space with that of the teacher model, leading to the loss of some probability flow. 

The necessity of further design details is discussed and experimentally validated in the Supplementary Materials.

\begin{table}
\begin{center}
\setlength{\tabcolsep}{3.6mm}
\caption{Quantitative Results on GSO of the Ablation Study in Fig.~\ref{fig7}.}
\vspace{-3pt}
\label{tab3}
\begin{tabular}{cccc}
    \toprule
    Objective & PSNR $\uparrow$ & SSIM $\uparrow$ & LPIPS $\downarrow$  \\
    \midrule
    Explicit Supervision & 19.06 &  0.862 &  0.147   \\
    Implicit Verification & 18.47 &  0.849 &  0.160  \\ \midrule
    Joint Optimization & \textbf{19.85} & \textbf{0.883} & \textbf{0.131}  \\
    \bottomrule
\end{tabular}
\end{center}
\vspace{-8pt}
\end{table}

\begin{figure}[t]
  \centering
   \includegraphics[width=0.96\linewidth]{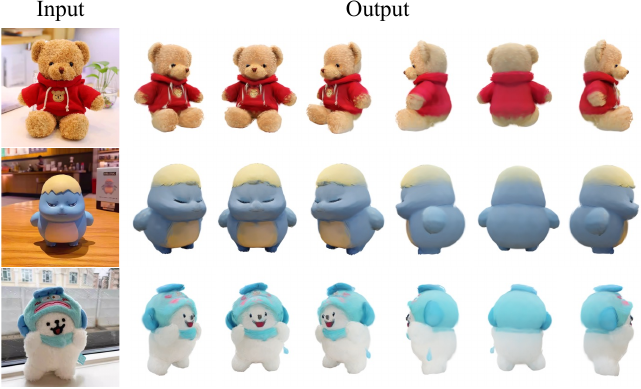}
   \vspace{-3pt}
   \caption{Samples generated from photographs taken with mobile phones.}
   \label{fig8}
\end{figure}

\subsection{Experiments on Photographs}
To evaluate the applicability and robustness of our proposed method in realistic scenarios, we perform additional experiments utilizing real-world photographs captured by smartphone cameras, as presented in Fig.~\ref{fig8}. Remarkably, even though our training dataset does not include photographic images, our model successfully generalizes and produces high-quality 3D reconstructions from these images. The synthesized results demonstrate accurate geometry, underscoring our method's strong ability to handle data variations inherent in real-world photography, such as varying illumination, viewpoint changes, background complexity, and slight noise. These results validate the practical value and versatility of our method, suggesting its potential effectiveness in broader real-life applications, particularly in scenarios involving data captured by consumer-level devices.

\begin{table}[t]
\begin{center}
    \setlength{\tabcolsep}{2.3mm}
    \caption{Quantitative Experimental Results on GSO Obtained by Models Distilled from Different Training Datasets.}
    \label{tab4}
    \begin{tabular}{cccc}
    \toprule
    Type and Quantity & PSNR $\uparrow$ & SSIM $\uparrow$ & LPIPS $\downarrow$  \\
    \midrule
    Rendered (40k) & 18.32 & 0.830 & 0.164  \\ 
    Synthetic (40k) & 16.55 & 0.807 & 0.203 \\
    Rendered (20k) + Synthetic (40k) & 18.36 & 0.831 & 0.158 \\ \midrule
    Rendered (40k) + Synthetic (80k)& \textbf{19.85} & \textbf{0.883} & \textbf{0.131}  \\
    \bottomrule
    \end{tabular}
    \label{tab4}
\end{center}
\vspace{-5pt}
\end{table}

\begin{figure*}[t]
  \centering
   \includegraphics[width=\linewidth]{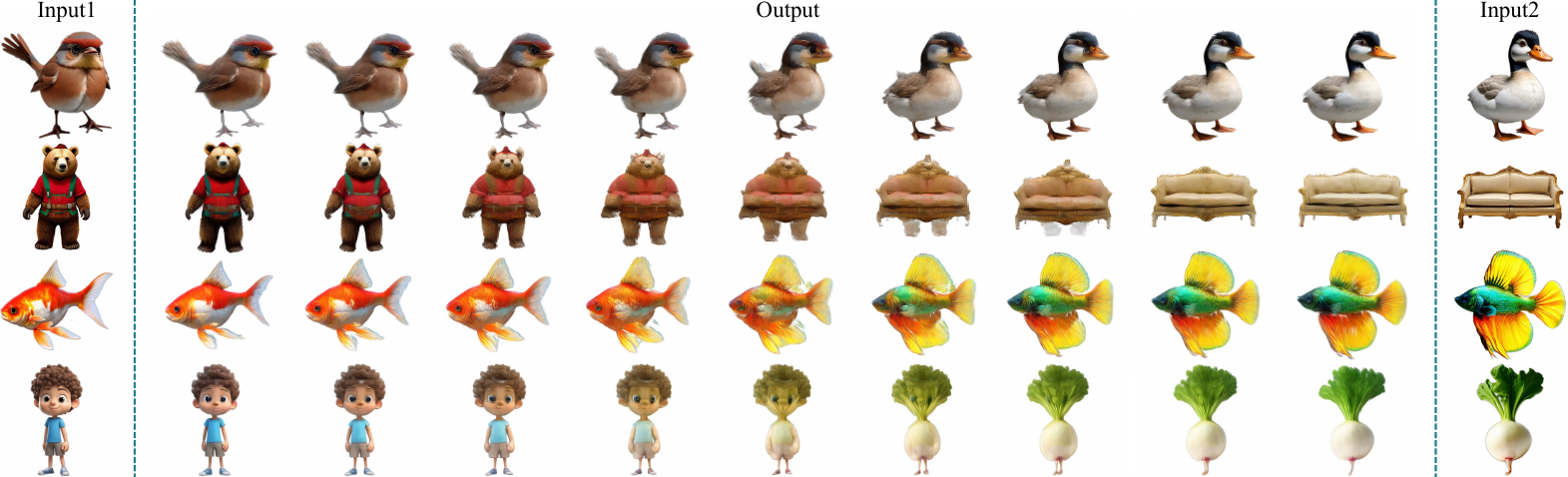}
   \caption{Illustration of PEPD's output when using interpolated input embeddings $(1-\beta)II_1 + \beta II_2$.}
   \vspace{-5pt}
   \label{fig9}
\end{figure*}

\subsection{Ablation Study of the Dataset} 

We collect a comprehensive dataset comprising 120k RGBA images to facilitate effective knowledge distillation from MV-DM. The dataset includes a diverse combination of synthetic and rendered images, aiming to provide sufficient variability and complexity for robust model training. To systematically evaluate the impact of dataset composition on the performance of the distilled PEPD model, we conduct a series of ablation experiments. As detailed in Tab.~\ref{tab4}, we methodically vary the type and quantity of synthetic and rendered images and subsequently evaluate the trained models on the GSO benchmark. Our results indicate that: 1) increasing the total volume of training data consistently improves model accuracy, underscoring the model's ability to effectively leverage larger datasets; and 2) perhaps more notably, rendered images exhibit a disproportionately positive impact on performance compared to synthetic images, likely because their distribution more closely resembles that of the test data.

\subsection{Interpolation Results}

To further validate the generalization ability of PEPD after distillation training, we conduct a latent space interpolation experiment. We fix \textit{N} and select two distinct input images, extracting their latent features via DINO encoding \cite{oquab2023dinov2}. A linear interpolation between these representations yields intermediate codes that we feed into PEPD to generate corresponding 3D outputs. As shown in Fig.~\ref{fig9}, the outputs transition smoothly as the interpolation weights vary. This smooth transformation indicates that our model learns a continuous latent space capturing meaningful geometric variations rather than merely memorizing the training set. The consistent generative process confirms that our distillation training effectively transfers knowledge, enhancing robustness and generalization to unseen inputs, and underscores the importance of a well-structured latent space for 3D reconstruction.

\begin{figure}[t]
  \centering
   \includegraphics[width=0.9\linewidth]{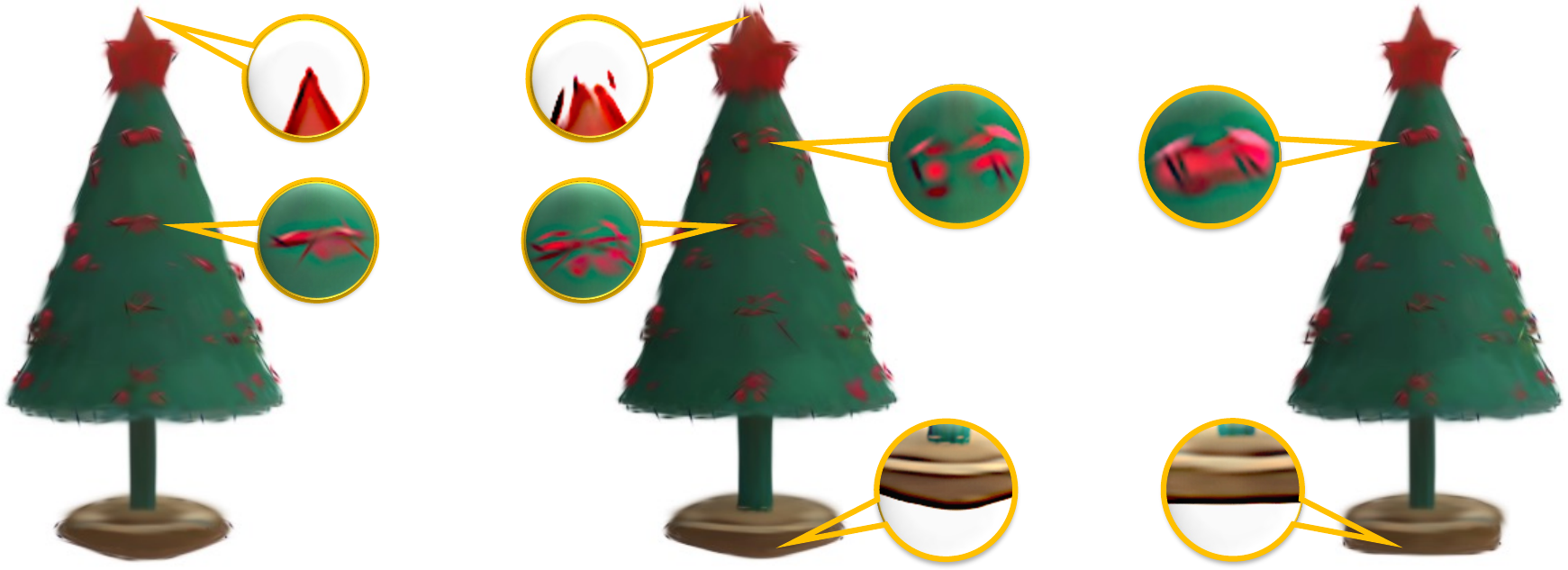}
   \caption{Illustration of multiple inferences performed on the same Christmas tree image. While there are certain differences in texture and structural details among the multiple inferences, the overall layout of the various elements shows minimal variation when compared to the 2D diffusion model.}
   \vspace{-5pt}
   \label{fig10}
\end{figure}

\subsection{Discussion and Prospects}
\label{sec:4.4}

We perform inference on a Christmas tree image using the trained PEPD model and render the results of three inference runs from the same viewpoint, as shown in Fig.~\ref{fig10}. The multiple inference results exhibit some variation in texture and structure. However, we observe that the model tends to generate in a conservative manner (e.g., it rarely produces large deformations, even in occluded regions), resulting in reduced sample diversity compared to the original MV-DM. We attribute this to the insufficient quantity of RGBA images and quadruples collected, which likely limits the full potential of DD3G, placing the entire network in a state of ``under-distillation.'' In addition, the reduction in diversity is also a challenge in the 2D distillation domain \cite{kang2024distilling}, suggesting that, beyond increasing the dataset size, further research into the distillation mechanism is necessary.

To strike a balance between computational cost and performance, we select ImageDream \cite{wang2023imagedream} as the teacher model and verifier. It is foreseeable that using larger models with more views, such as SV3D \cite{voleti2025sv3d}, could further enhance the distillation process. However, this will lead to a significant increase in computational demands, estimated at around 40,000 A100 GPU hours. Fortunately, collecting more RGBA image data and increasing computational power is relatively straightforward for the broader research community. DD3G holds substantial potential, as it no longer requires the involvement of 3D data during  the distillation process.

\section{Conclusion}

In this paper, we propose DD3G, a formulation that enables the distillation of knowledge from MV-DM into a 3D generator. To align the student model with the teacher model in the representation space and achieve efficient 3D Gaussian generation in weakly-supervised settings, we design a 3D Gaussian generator, PEPD, consisting of two phases: Pattern Extraction and Progressive Decoding. Moreover, by combining explicit supervision with implicit verification objectives, we perform joint optimization for PEPD, mitigating issues such as knowledge loss during the distillation process and unreasonable views. Qualitative and quantitative experiments demonstrate the effectiveness of our method. Since 3D data is no longer required during the distillation process, DD3G is expected to achieve even better results in the future as the scale of RGBA image data increases.

\bibliography{main}

\begin{thebibliography}{10}
\providecommand{\url}[1]{#1}
\csname url@samestyle\endcsname
\providecommand{\newblock}{\relax}
\providecommand{\bibinfo}[2]{#2}
\providecommand{\BIBentrySTDinterwordspacing}{\spaceskip=0pt\relax}
\providecommand{\BIBentryALTinterwordstretchfactor}{4}
\providecommand{\BIBentryALTinterwordspacing}{\spaceskip=\fontdimen2\font plus
\BIBentryALTinterwordstretchfactor\fontdimen3\font minus
  \fontdimen4\font\relax}
\providecommand{\BIBforeignlanguage}[2]{{%
\expandafter\ifx\csname l@#1\endcsname\relax
\typeout{** WARNING: IEEEtran.bst: No hyphenation pattern has been}%
\typeout{** loaded for the language `#1'. Using the pattern for}%
\typeout{** the default language instead.}%
\else
\language=\csname l@#1\endcsname
\fi
#2}}
\providecommand{\BIBdecl}{\relax}
\BIBdecl

\bibitem{rombach2022high}
R.~Rombach, A.~Blattmann, D.~Lorenz, P.~Esser, and B.~Ommer, ``High-resolution
  image synthesis with latent diffusion models,'' in \emph{Proceedings of the
  IEEE/CVF conference on computer vision and pattern recognition}, 2022, pp.
  10\,684--10\,695.

\bibitem{kerbl20233d}
B.~Kerbl, G.~Kopanas, T.~Leimk{\"u}hler, and G.~Drettakis, ``3d gaussian
  splatting for real-time radiance field rendering.'' \emph{ACM Trans. Graph.},
  vol.~42, no.~4, pp. 139--1, 2023.

\bibitem{huang2025learning}
S.-Y. Huang, C.-P. Huang, K.-P. Chang, Z.-T. Chou, I.-J. Liu, and Y.-C.~F.
  Wang, ``Learning shape-color diffusion priors for text-guided 3d object
  generation,'' \emph{IEEE Transactions on Multimedia}, 2025.

\bibitem{wang2023imagedream}
P.~Wang and Y.~Shi, ``Imagedream: Image-prompt multi-view diffusion for 3d
  generation,'' \emph{arXiv preprint arXiv:2312.02201}, 2023.

\bibitem{voleti2025sv3d}
V.~Voleti, C.-H. Yao, M.~Boss, A.~Letts, D.~Pankratz, D.~Tochilkin, C.~Laforte,
  R.~Rombach, and V.~Jampani, ``Sv3d: Novel multi-view synthesis and 3d
  generation from a single image using latent video diffusion,'' in
  \emph{European Conference on Computer Vision}.\hskip 1em plus 0.5em minus
  0.4em\relax Springer, 2025, pp. 439--457.

\bibitem{poole2022dreamfusion}
B.~Poole, A.~Jain, J.~T. Barron, and B.~Mildenhall, ``Dreamfusion: Text-to-3d
  using 2d diffusion,'' \emph{arXiv preprint arXiv:2209.14988}, 2022.

\bibitem{tang2023dreamgaussian}
J.~Tang, J.~Ren, H.~Zhou, Z.~Liu, and G.~Zeng, ``Dreamgaussian: Generative
  gaussian splatting for efficient 3d content creation,'' \emph{arXiv preprint
  arXiv:2309.16653}, 2023.

\bibitem{yi2024gaussiandreamer}
T.~Yi, J.~Fang, J.~Wang, G.~Wu, L.~Xie, X.~Zhang, W.~Liu, Q.~Tian, and X.~Wang,
  ``Gaussiandreamer: Fast generation from text to 3d gaussians by bridging 2d
  and 3d diffusion models,'' in \emph{Proceedings of the IEEE/CVF Conference on
  Computer Vision and Pattern Recognition}, 2024, pp. 6796--6807.

\bibitem{li2024instant3d}
M.~Li, P.~Zhou, J.-W. Liu, J.~Keppo, M.~Lin, S.~Yan, and X.~Xu, ``Instant3d:
  Instant text-to-3d generation,'' \emph{International Journal of Computer
  Vision}, pp. 1--17, 2024.

\bibitem{tang2025lgm}
J.~Tang, Z.~Chen, X.~Chen, T.~Wang, G.~Zeng, and Z.~Liu, ``Lgm: Large
  multi-view gaussian model for high-resolution 3d content creation,'' in
  \emph{European Conference on Computer Vision}.\hskip 1em plus 0.5em minus
  0.4em\relax Springer, 2025, pp. 1--18.

\bibitem{amos2022tutorial}
B.~Amos, ``Tutorial on amortized optimization for learning to optimize over
  continuous domains,'' \emph{arXiv preprint arXiv:2202.00665}, vol.~2, no.~3,
  p.~9, 2022.

\bibitem{xie2024latte3d}
K.~Xie, J.~Lorraine, T.~Cao, J.~Gao, J.~Lucas, A.~Torralba, S.~Fidler, and
  X.~Zeng, ``Latte3d: Large-scale amortized text-to-enhanced3d synthesis,''
  \emph{arXiv preprint arXiv:2403.15385}, 2024.

\bibitem{song2020denoising}
J.~Song, C.~Meng, and S.~Ermon, ``Denoising diffusion implicit models,''
  \emph{arXiv preprint arXiv:2010.02502}, 2020.

\bibitem{lin2024sdxl}
S.~Lin, A.~Wang, and X.~Yang, ``Sdxl-lightning: Progressive adversarial
  diffusion distillation,'' \emph{arXiv preprint arXiv:2402.13929}, 2024.

\bibitem{kang2024distilling}
M.~Kang, R.~Zhang, C.~Barnes, S.~Paris, S.~Kwak, J.~Park, E.~Shechtman, J.-Y.
  Zhu, and T.~Park, ``Distilling diffusion models into conditional gans,''
  \emph{arXiv preprint arXiv:2405.05967}, 2024.

\bibitem{yin2024one}
T.~Yin, M.~Gharbi, R.~Zhang, E.~Shechtman, F.~Durand, W.~T. Freeman, and
  T.~Park, ``One-step diffusion with distribution matching distillation,'' in
  \emph{Proceedings of the IEEE/CVF Conference on Computer Vision and Pattern
  Recognition}, 2024, pp. 6613--6623.

\bibitem{achiam2023gpt}
J.~Achiam, S.~Adler, S.~Agarwal, L.~Ahmad, I.~Akkaya, F.~L. Aleman, D.~Almeida,
  J.~Altenschmidt, S.~Altman, S.~Anadkat \emph{et~al.}, ``Gpt-4 technical
  report,'' \emph{arXiv preprint arXiv:2303.08774}, 2023.

\bibitem{esser2024scaling}
P.~Esser, S.~Kulal, A.~Blattmann, R.~Entezari, J.~M{\"u}ller, H.~Saini,
  Y.~Levi, D.~Lorenz, A.~Sauer, F.~Boesel \emph{et~al.}, ``Scaling rectified
  flow transformers for high-resolution image synthesis,'' in \emph{Forty-first
  International Conference on Machine Learning}, 2024.

\bibitem{kirillov2023segment}
A.~Kirillov, E.~Mintun, N.~Ravi, H.~Mao, C.~Rolland, L.~Gustafson, T.~Xiao,
  S.~Whitehead, A.~C. Berg, W.-Y. Lo \emph{et~al.}, ``Segment anything,'' in
  \emph{Proceedings of the IEEE/CVF International Conference on Computer
  Vision}, 2023, pp. 4015--4026.

\bibitem{deitke2023objaverse}
M.~Deitke, D.~Schwenk, J.~Salvador, L.~Weihs, O.~Michel, E.~VanderBilt,
  L.~Schmidt, K.~Ehsani, A.~Kembhavi, and A.~Farhadi, ``Objaverse: A universe
  of annotated 3d objects,'' in \emph{Proceedings of the IEEE/CVF Conference on
  Computer Vision and Pattern Recognition}, 2023, pp. 13\,142--13\,153.

\bibitem{wang2023score}
H.~Wang, X.~Du, J.~Li, R.~A. Yeh, and G.~Shakhnarovich, ``Score jacobian
  chaining: Lifting pretrained 2d diffusion models for 3d generation,'' in
  \emph{Proceedings of the IEEE/CVF Conference on Computer Vision and Pattern
  Recognition}, 2023, pp. 12\,619--12\,629.

\bibitem{ma2025scaledreamer}
Z.~Ma, Y.~Wei, Y.~Zhang, X.~Zhu, Z.~Lei, and L.~Zhang, ``Scaledreamer: Scalable
  text-to-3d synthesis with asynchronous score distillation,'' in
  \emph{European Conference on Computer Vision}.\hskip 1em plus 0.5em minus
  0.4em\relax Springer, 2025, pp. 1--19.

\bibitem{wang2024prolificdreamer}
Z.~Wang, C.~Lu, Y.~Wang, F.~Bao, C.~Li, H.~Su, and J.~Zhu, ``Prolificdreamer:
  High-fidelity and diverse text-to-3d generation with variational score
  distillation,'' \emph{Advances in Neural Information Processing Systems},
  vol.~36, 2024.

\bibitem{zhuo2024vividdreamer}
W.~Zhuo, F.~Ma, H.~Fan, and Y.~Yang, ``Vividdreamer: Invariant score
  distillation for hyper-realistic text-to-3d generation,'' \emph{arXiv
  preprint arXiv:2407.09822}, 2024.

\bibitem{xu2019disn}
Q.~Xu, W.~Wang, D.~Ceylan, R.~Mech, and U.~Neumann, ``Disn: Deep implicit
  surface network for high-quality single-view 3d reconstruction,''
  \emph{Advances in neural information processing systems}, vol.~32, 2019.

\bibitem{mescheder2019occupancy}
L.~Mescheder, M.~Oechsle, M.~Niemeyer, S.~Nowozin, and A.~Geiger, ``Occupancy
  networks: Learning 3d reconstruction in function space,'' in
  \emph{Proceedings of the IEEE/CVF conference on computer vision and pattern
  recognition}, 2019, pp. 4460--4470.

\bibitem{xu2023dmv3d}
Y.~Xu, H.~Tan, F.~Luan, S.~Bi, P.~Wang, J.~Li, Z.~Shi, K.~Sunkavalli,
  G.~Wetzstein, Z.~Xu \emph{et~al.}, ``Dmv3d: Denoising multi-view diffusion
  using 3d large reconstruction model,'' \emph{arXiv preprint
  arXiv:2311.09217}, 2023.

\bibitem{kwon2025text2avatar}
Y.-H. Kwon, J.~H. Yoon, and M.-G. Park, ``Text2avatar: Articulated 3d avatar
  creation with text instructions,'' \emph{IEEE Transactions on Multimedia},
  2025.

\bibitem{yin2025shapegpt}
F.~Yin, X.~Chen, C.~Zhang, B.~Jiang, Z.~Zhao, W.~Liu, G.~Yu, and T.~Chen,
  ``Shapegpt: 3d shape generation with a unified multi-modal language model,''
  \emph{IEEE Transactions on Multimedia}, 2025.

\bibitem{hong2023lrm}
Y.~Hong, K.~Zhang, J.~Gu, S.~Bi, Y.~Zhou, D.~Liu, F.~Liu, K.~Sunkavalli,
  T.~Bui, and H.~Tan, ``Lrm: Large reconstruction model for single image to
  3d,'' \emph{arXiv preprint arXiv:2311.04400}, 2023.

\bibitem{han2025vfusion3d}
J.~Han, F.~Kokkinos, and P.~Torr, ``Vfusion3d: Learning scalable 3d generative
  models from video diffusion models,'' in \emph{European Conference on
  Computer Vision}.\hskip 1em plus 0.5em minus 0.4em\relax Springer, 2025, pp.
  333--350.

\bibitem{girdhar2023emu}
R.~Girdhar, M.~Singh, A.~Brown, Q.~Duval, S.~Azadi, S.~S. Rambhatla, A.~Shah,
  X.~Yin, D.~Parikh, and I.~Misra, ``Emu video: Factorizing text-to-video
  generation by explicit image conditioning,'' \emph{arXiv preprint
  arXiv:2311.10709}, 2023.

\bibitem{chen2024survey}
G.~Chen and W.~Wang, ``A survey on 3d gaussian splatting,'' \emph{arXiv
  preprint arXiv:2401.03890}, 2024.

\bibitem{yu2024mip}
Z.~Yu, A.~Chen, B.~Huang, T.~Sattler, and A.~Geiger, ``Mip-splatting:
  Alias-free 3d gaussian splatting,'' in \emph{Proceedings of the IEEE/CVF
  Conference on Computer Vision and Pattern Recognition}, 2024, pp.
  19\,447--19\,456.

\bibitem{zou2024triplane}
Z.-X. Zou, Z.~Yu, Y.-C. Guo, Y.~Li, D.~Liang, Y.-P. Cao, and S.-H. Zhang,
  ``Triplane meets gaussian splatting: Fast and generalizable single-view 3d
  reconstruction with transformers,'' in \emph{Proceedings of the IEEE/CVF
  Conference on Computer Vision and Pattern Recognition}, 2024, pp.
  10\,324--10\,335.

\bibitem{xu2024agg}
D.~Xu, Y.~Yuan, M.~Mardani, S.~Liu, J.~Song, Z.~Wang, and A.~Vahdat, ``Agg:
  Amortized generative 3d gaussians for single image to 3d,'' \emph{arXiv
  preprint arXiv:2401.04099}, 2024.

\bibitem{jiang2024brightdreamer}
L.~Jiang and L.~Wang, ``Brightdreamer: Generic 3d gaussian generative framework
  for fast text-to-3d synthesis,'' \emph{arXiv preprint arXiv:2403.11273},
  2024.

\bibitem{szymanowicz2024splatter}
S.~Szymanowicz, C.~Rupprecht, and A.~Vedaldi, ``Splatter image: Ultra-fast
  single-view 3d reconstruction,'' in \emph{Proceedings of the IEEE/CVF
  Conference on Computer Vision and Pattern Recognition}, 2024, pp.
  10\,208--10\,217.

\bibitem{he2025gvgen}
X.~He, J.~Chen, S.~Peng, D.~Huang, Y.~Li, X.~Huang, C.~Yuan, W.~Ouyang, and
  T.~He, ``Gvgen: Text-to-3d generation with volumetric representation,'' in
  \emph{European Conference on Computer Vision}.\hskip 1em plus 0.5em minus
  0.4em\relax Springer, 2025, pp. 463--479.

\bibitem{zhang2024gaussiancube}
B.~Zhang, Y.~Cheng, J.~Yang, C.~Wang, F.~Zhao, Y.~Tang, D.~Chen, and B.~Guo,
  ``Gaussiancube: Structuring gaussian splatting using optimal transport for 3d
  generative modeling,'' \emph{arXiv preprint arXiv:2403.19655}, 2024.

\bibitem{ho2020denoising}
J.~Ho, A.~Jain, and P.~Abbeel, ``Denoising diffusion probabilistic models,''
  \emph{Advances in neural information processing systems}, vol.~33, pp.
  6840--6851, 2020.

\bibitem{jiang2024gs}
Y.~Jiang, J.~Li, H.~Qin, Y.~Dai, J.~Liu, G.~Zhang, C.~Zhang, and T.~Yang,
  ``Gs-sfs: Joint gaussian splatting and shape-from-silhouette for multiple
  human reconstruction in large-scale sports scenes,'' \emph{IEEE Transactions
  on Multimedia}, 2024.

\bibitem{yi2024diffusion}
X.~Yi, Z.~Wu, Q.~Xu, P.~Zhou, J.-H. Lim, and H.~Zhang, ``Diffusion time-step
  curriculum for one image to 3d generation,'' in \emph{Proceedings of the
  IEEE/CVF Conference on Computer Vision and Pattern Recognition}, 2024, pp.
  9948--9958.

\bibitem{shi2023mvdream}
Y.~Shi, P.~Wang, J.~Ye, M.~Long, K.~Li, and X.~Yang, ``Mvdream: Multi-view
  diffusion for 3d generation,'' \emph{arXiv preprint arXiv:2308.16512}, 2023.

\bibitem{wang2019dgl}
M.~Wang, D.~Zheng, Z.~Ye, Q.~Gan, M.~Li, X.~Song, J.~Zhou, C.~Ma, L.~Yu,
  Y.~Gai, T.~Xiao, T.~He, G.~Karypis, J.~Li, and Z.~Zhang, ``Deep graph
  library: A graph-centric, highly-performant package for graph neural
  networks,'' \emph{arXiv preprint arXiv:1909.01315}, 2019.

\bibitem{zhang2018unreasonable}
R.~Zhang, P.~Isola, A.~A. Efros, E.~Shechtman, and O.~Wang, ``The unreasonable
  effectiveness of deep features as a perceptual metric,'' in \emph{Proceedings
  of the IEEE conference on computer vision and pattern recognition}, 2018, pp.
  586--595.

\bibitem{chen2024v3d}
Z.~Chen, Y.~Wang, F.~Wang, Z.~Wang, and H.~Liu, ``V3d: Video diffusion models
  are effective 3d generators,'' \emph{arXiv preprint arXiv:2403.06738}, 2024.

\bibitem{downs2022google}
L.~Downs, A.~Francis, N.~Koenig, B.~Kinman, R.~Hickman, K.~Reymann, T.~B.
  McHugh, and V.~Vanhoucke, ``Google scanned objects: A high-quality dataset of
  3d scanned household items,'' in \emph{2022 International Conference on
  Robotics and Automation (ICRA)}.\hskip 1em plus 0.5em minus 0.4em\relax IEEE,
  2022, pp. 2553--2560.

\bibitem{tremblay2022rtmv}
J.~Tremblay, M.~Meshry, A.~Evans, J.~Kautz, A.~Keller, S.~Khamis,
  T.~M{\"u}ller, C.~Loop, N.~Morrical, K.~Nagano \emph{et~al.}, ``Rtmv: A
  ray-traced multi-view synthetic dataset for novel view synthesis,''
  \emph{arXiv preprint arXiv:2205.07058}, 2022.

\bibitem{kingma2014adam}
D.~P. Kingma and J.~Ba, ``Adam: A method for stochastic optimization,''
  \emph{arXiv preprint arXiv:1412.6980}, 2014.

\bibitem{radford2021learning}
A.~Radford, J.~W. Kim, C.~Hallacy, A.~Ramesh, G.~Goh, S.~Agarwal, G.~Sastry,
  A.~Askell, P.~Mishkin, J.~Clark \emph{et~al.}, ``Learning transferable visual
  models from natural language supervision,'' in \emph{International conference
  on machine learning}.\hskip 1em plus 0.5em minus 0.4em\relax PMLR, 2021, pp.
  8748--8763.

\bibitem{oquab2023dinov2}
M.~Oquab, T.~Darcet, T.~Moutakanni, H.~Vo, M.~Szafraniec, V.~Khalidov,
  P.~Fernandez, D.~Haziza, F.~Massa, A.~El-Nouby \emph{et~al.}, ``Dinov2:
  Learning robust visual features without supervision,'' \emph{arXiv preprint
  arXiv:2304.07193}, 2023.

\end{thebibliography}

\vfill

\end{document}


\title{Appendix for DD3G}

\author{Hao Qin, Luyuan Chen, Ming Kong\textsuperscript{*}, Mengxu Lu, Qiang Zhu
	\thanks{\textsuperscript{*} Corresponding author.}
	\thanks{Hao Qin, Ming Kong, Mengxu Lu, and Qiang Zhu are with School of Computer Science and Technology, Zhejiang University,  Hangzhou 310027, China (e-mail: haoqin@zju.edu.cn; zjukongming@zju.edu.cn; lumengxu@zju.edu.cn; zhuq@zju.edu.cn).}
	\thanks{Luyuan Chen is with School of Computer, Beijing Information Science and Technology University, Beijing 100005, China (email: chenly@bistu.edu.cn).}}

\markboth{Journal of \LaTeX\ Class Files,~Vol.~14, No.~8, August~2021}%
{Shell \MakeLowercase{\textit{et al.}}: A Sample Article Using IEEEtran.cls for IEEE Journals}

\maketitle

\begin{figure}[t]
  \centering
   \includegraphics[width=\linewidth]{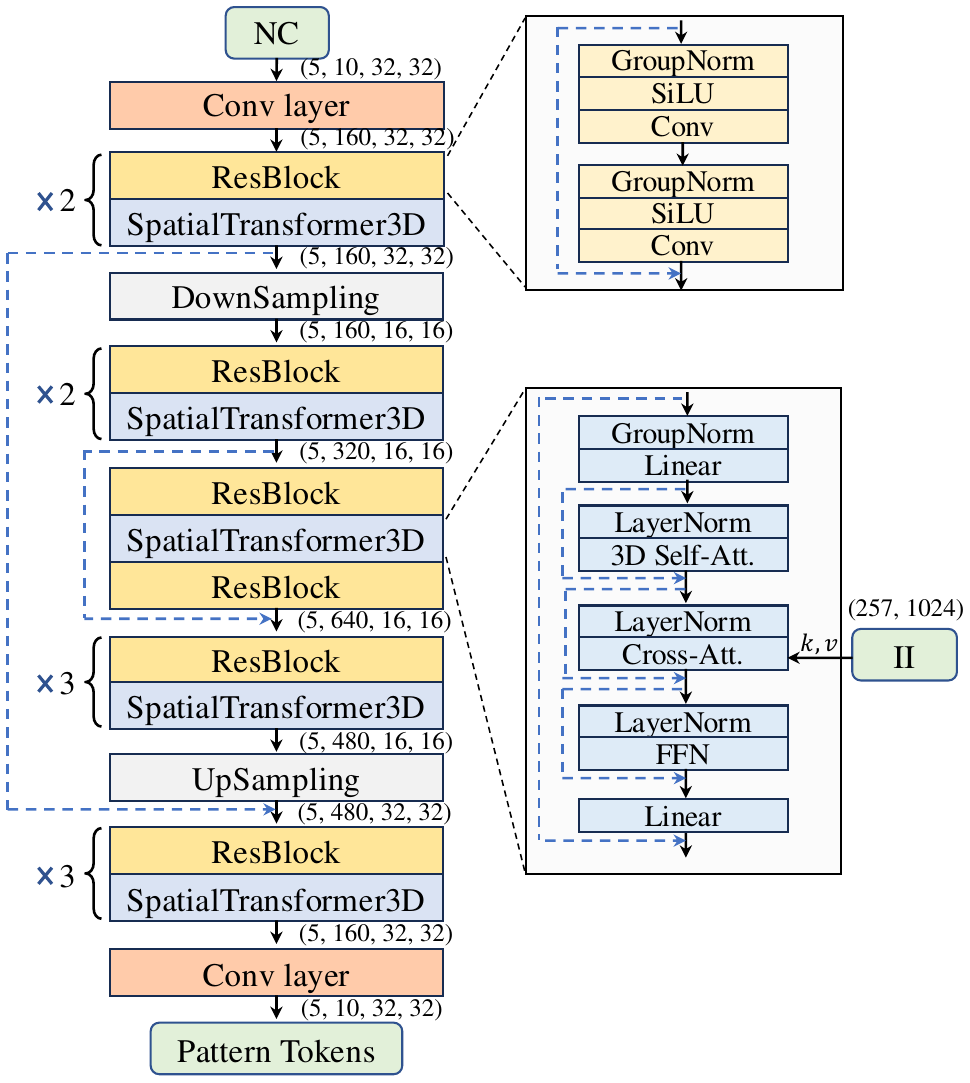}
   \caption{Specific structure of Pattern Extraction. The blue dotted line indicates the residual connection}
   \label{figs1}
\end{figure}
\begin{figure}[t]
  \centering
   \includegraphics[width=\linewidth]{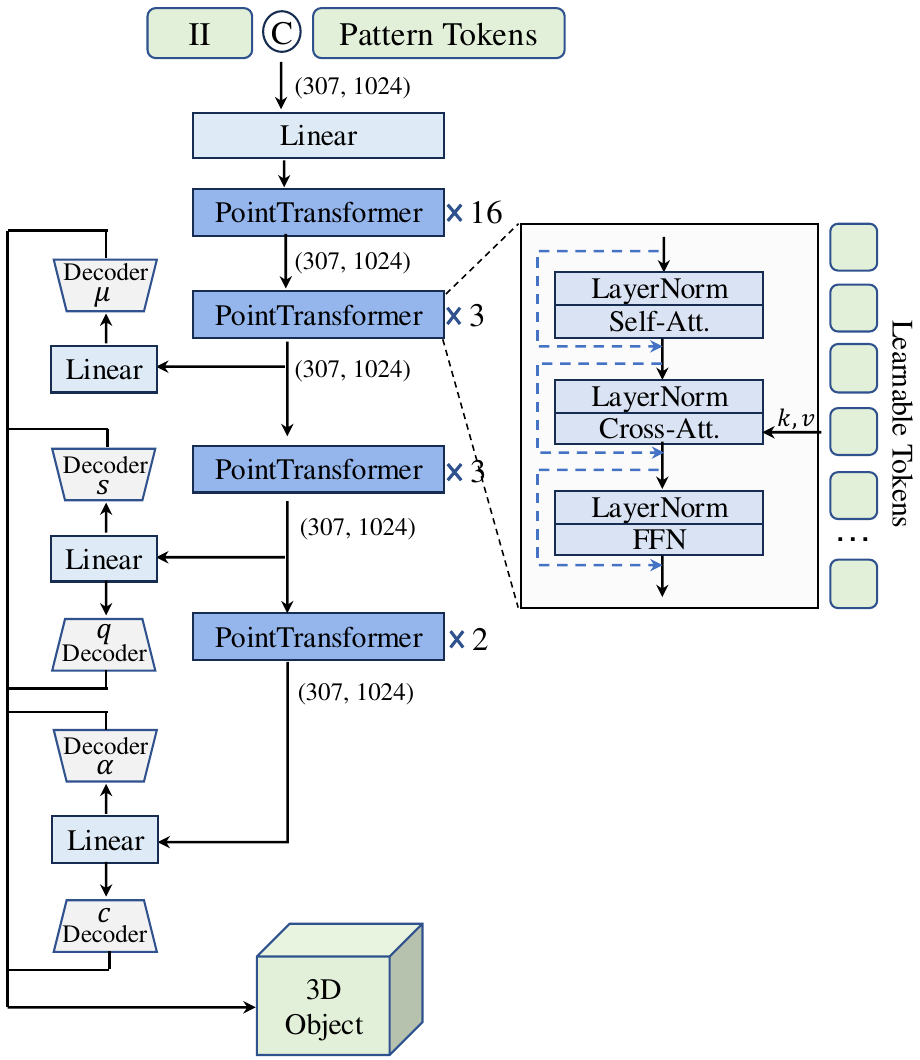}
   \caption{Specific structure of Progressive Decoding.}
   \label{figs2}
\end{figure}

\section{Overview}
Here, we provide additional implementation details and experimental results as a supplement to the main text. The specific contents of the appendix are as follows: Sec.~\ref{S2} provides a detailed description and demonstration of the specific structure of PEPD; Sec.~\ref{S3} offers comparisons with state-of-the-art Triplane and Mesh generation methods; Sec.~\ref{S4} presents additional ablation studies on PEPD to validate the effectiveness of Pattern Extraction and Progressive Decoding; Sec.~\ref{S5} displays some specifics of the synthetic RGBA image collection process;  and Sec.~\ref{S7} demonstrates an ablation study of random background colors.

\section{Architecture of PEPD}
\label{S2}
In the main text, we briefly introduce the basic components used in PEPD; here, we provide a detailed description of the specific structure of PEPD.
Let the input image be denoted as  $II_i \in \mathbb{R}^{3*256*256}$, which we encode using DINOv2 \cite{oquab2023dinov2} to obtain its feature $F_i \in \mathbb{R}^{257*1024}$. Assume the noise sampled from the Gaussian distribution is $N_i$, and since our teacher model is ImageDream \cite{wang2023imagedream}, $N_i \in \mathbb{R}^{5*4*32*32}$. In ImageDream, the initial camera pose $C_i$ is represented by a matrix of dimensions \( 5 \times 4 \times 4 \). After encoding through Plücker ray embedding and a dimension transformation, it is converted into a matrix of dimensions \( 5 \times 6 \times 32 \times 32 \), i.e., $C_i \in \mathbb{R}^{5*6*32*32}$. $N_i$ and $C_i$ are concatenated along the channel dimension to form the input $NC_i \in \mathbb{R}^{5*10*32*32}$ for Pattern Extraction.

\textbf{Pattern Extraction (PE):} We aim to infuse the information from $II_i$ into $NC_i$ and facilitate certain interactions among different sub-random variables within $NC_i$. Some existing multi-view U-Nets have already achieved this purpose \cite{tang2025lgm, shi2023mvdream, wang2023imagedream}, however, in our experiments, we find that using existing structures tends to lead the model into local optima. We speculate that this issue mainly arises from the conflict between the large number of parameters, the indirect optimization objectives, and the relatively insufficient training data. Considering our desire in PE to provide a generalized description of the image's lifting pattern without needing to extract detailed information, we design a more streamlined mapper. This involves reducing the frequent upsampling and downsampling operations found in the original U-Net and decreasing model complexity. The specific structure of PE is shown in Fig.~\ref{figs1}, where key intermediate feature dimensions are annotated.

\textbf{Progressive Decoding (PD):} Suppose the pattern tokens output from PE are denoted as $p_i \in \mathbb{R}^{5*10*32*32}$. After transforming the dimensions of $p_i$, we concatenate them with $F_i$ to serve as the input for PD. PD is composed of stacked point transformer layers, and its specific structure is shown in Fig.~\ref{figs2}.

\begin{table*}[t]
  \centering
    \setlength{\tabcolsep}{3.4mm}{
    
     \caption{Quantitative Comparison Between DD3G and Other Methods. ``w/o MV-DM'' Indicates that MV-DM is not Required During the Generation Process.}
    \label{tabs1}
    
    \begin{tabular}{c|c|ccc|ccc|c}
    \toprule
    \multirow{2}{*}{Method} & \multirow{2}{*}{w/o MV-DM} & \multicolumn{3}{c|}{Google Scanned Objects} & \multicolumn{3}{c|}{RTMV-bricks} & \multirow{2}{*}{Generation Time (s) $\downarrow$} \\
    & & PSNR $\uparrow$ & SSIM $\uparrow$ & LPIPS $\downarrow$ & PSNR $\uparrow$ & SSIM $\uparrow$ & LPIPS $\downarrow$ & \\ \midrule
    OpenLRM \cite{openlrm} & \ding{51} & 15.07 & 0.803 & 0.196 & 11.77 &  0.563 &  0.301 & \underline{0.23}  \\
    VFusion3D \cite{han2025vfusion3d} & \ding{51} & \cellcolor{red!0}18.92 & \cellcolor{red!0}0.871 & \underline{0.136} & 14.75 &  \underline{0.679} & 0.192 & \underline{0.23}  \\ \midrule
    One2345 \cite{liu2024one} & \ding{55} & 13.85 & 0.790 & 0.251 & 10.13 &  0.570 &  0.325 & 42  \\
    CRM \cite{wang2025crm} & \ding{55} & 16.96 & 0.846 & 0.165 & 12.03 & 0.624 & 0.273  & 10  \\ 
    InstantMesh \cite{xu2024instantmesh} & \ding{55} & \textbf{20.02} &  \underline{0.872} & 0.144 & \underline{15.62} & 0.675 &  \underline{0.180} & 7  \\ \midrule
    Ours & \ding{51} & \underline{19.85} & \textbf{0.883} & \textbf{0.131} & \textbf{15.83} &  \textbf{0.702} & \textbf{0.168} & \textbf{0.06}  \\
    \bottomrule
    \end{tabular}}

\end{table*}

\begin{figure}[t]
  \centering
   \includegraphics[width=\linewidth]{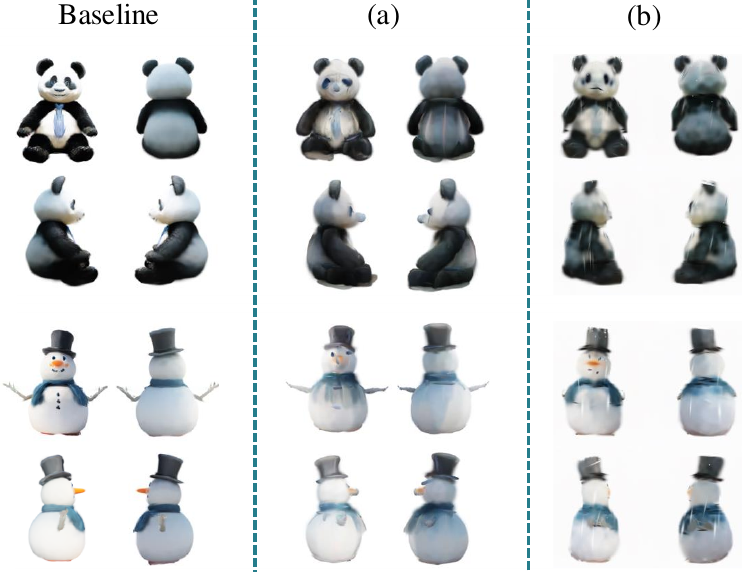}
   \caption{Ablation experiments of PEPD. ``Baseline" refers to the experimental results obtained when using the complete PEPD.}
   \label{figs3}
\end{figure}

\section{Comparison with Other Methods}
\label{S3}

In the main text, we limit our comparisons to 3D Gaussian generation methods. Here, we extend these comparisons to other representation techniques such as Triplane and Mesh generation methods on GSO \cite{downs2022google} and RTMV-bricks \cite{tremblay2022rtmv}. The experimental results are presented in Tab.~\ref{tabs1}. It is evident from the results that our method achieves a significant improvement in generation speed while maintaining a quality comparable to the current best-performing methods.

\section{More Ablation Experiments}
\label{S4}
To further illustrate the necessity of the detailed design in PEPD, we conduct additional ablation experiments.

\textbf{PE:} In PE, we design an approach to capture the lifting patterns from the input image. Initially, we intend to concatenate the random variable $NC$ directly with the input image $II$ to introduce probability flow; however, we find that this approach led to an unstable training process. To address this, we initially conduct multiple training rounds using only $II$ as the input for PD. Subsequently, we concatenate $NC$ with $II$ for training. This procedure enhance the stability of the training process, allowing the model to successfully converge. Experimental results, as shown in Fig.~\ref{figs3}a, indicate that in the absence of a preliminary extraction of the lifting pattern, the color boundaries in the generated samples become indistinct, with overlapping occurring between different regions.

\begin{table}[t]
    \centering
    
    \setlength{\tabcolsep}{3mm}{
    \caption{Quantitative Results on GSO of the Ablation Study in Fig.~\ref{figs3}.}
    \label{tabs2}
    \begin{tabular}{cccc}
    \toprule
    \textbf{Group} & \textbf{PSNR} $\uparrow$ & \textbf{SSIM} $\uparrow$ & \textbf{LPIPS} $\downarrow$  \\
    \midrule
    Baseline  & 19.85 & 0.883 & 0.131  \\
    (a) & 18.74 &  0.872 & 0.155   \\
    (b) & 17.08  & 0.859  &  0.182  \\
    \bottomrule
    \end{tabular}}
\end{table}

\textbf{PD:} We abandon the progressive decoding strategy in PD and instead decode all attributes of the 3D Gaussian through the features of final layer. The experimental results, as shown in Fig.~\ref{figs3}b, indicate severe detail loss and the emergence of numerous artifacts. This underscores the effectiveness of the progressive decoding strategy.

The quantitative results on GSO for the two groups of ablation experiments are presented in Tab.~\ref{tabs2}.

\begin{figure}[t]
  \centering
   \includegraphics[width=\linewidth]{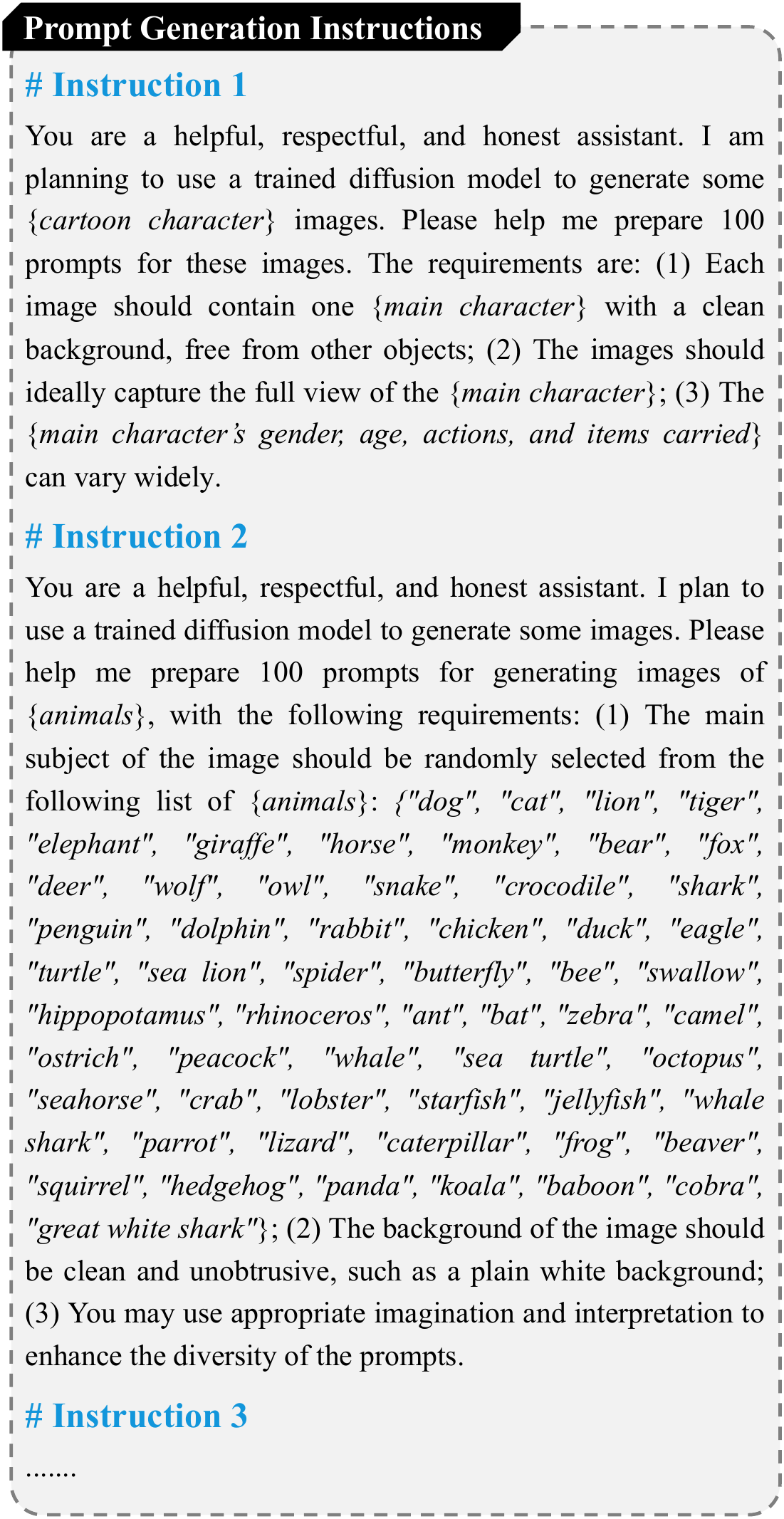}
   \caption{Instructions used to prompt GPT-4o.}
   \label{figs4}
\end{figure}

\begin{figure}[t]
  \centering
   \includegraphics[width=\linewidth]{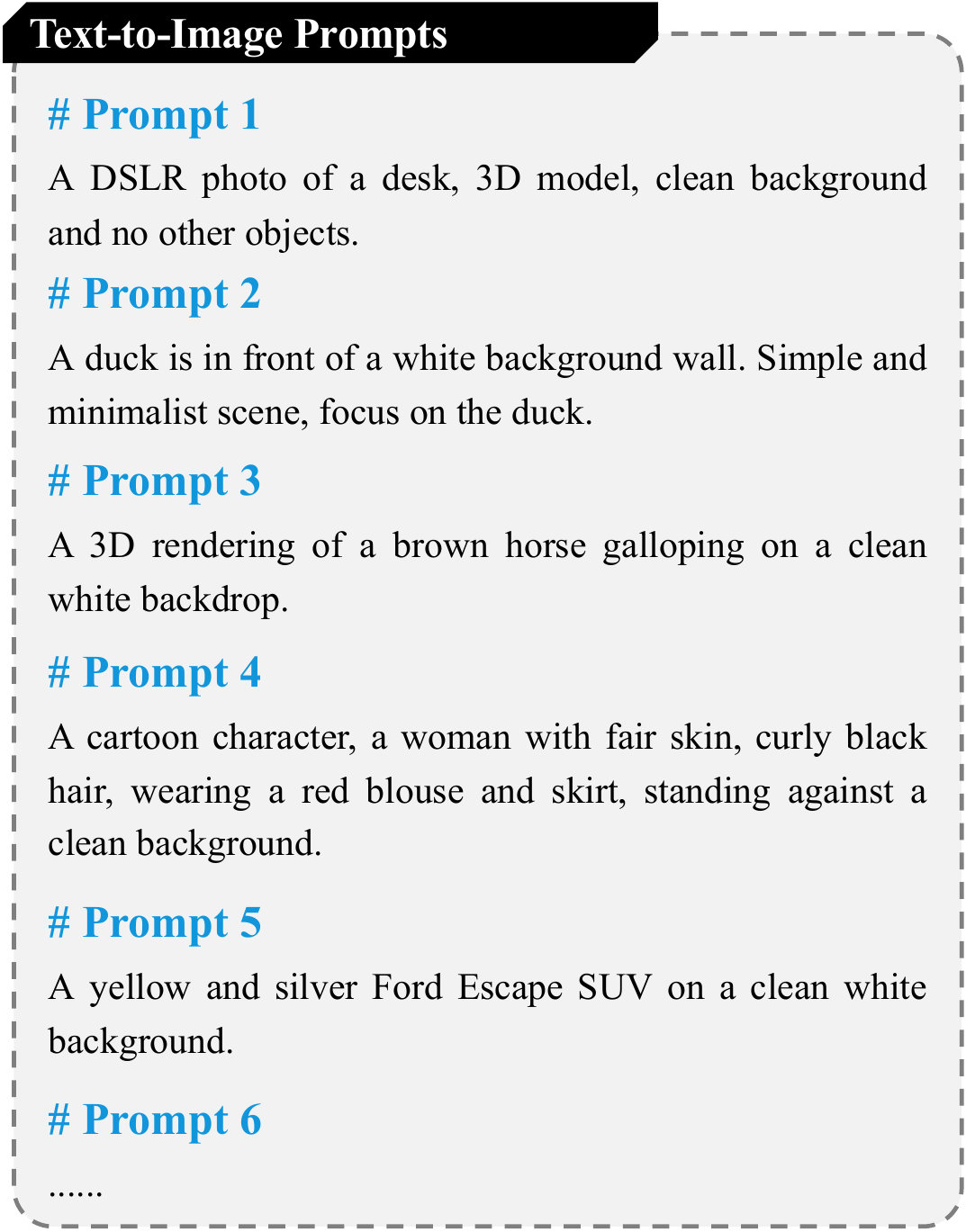}
   \caption{Examples of text-to-image prompts.}
   \label{figs5}
\end{figure}

\begin{figure*}[t]
  \centering
   \includegraphics[width=\linewidth]{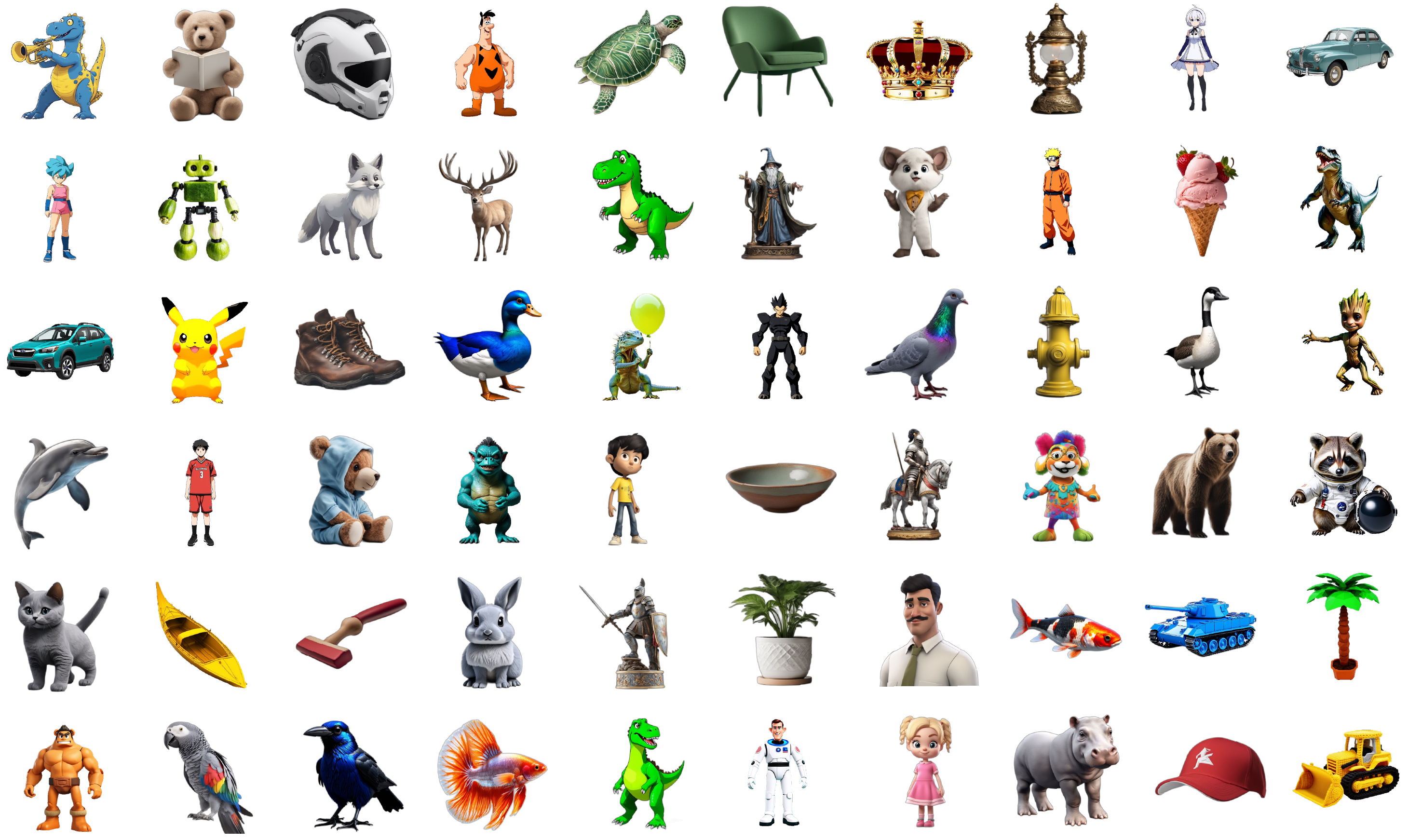}
   \caption{Some samples of synthetic RGBA images.}
   \label{figs6}
\end{figure*}

\section{More Details of Training Data}
\label{S5}
To train PEPD, we synthesize 80,945 RGBA images. The synthesis process includes the following stages: Prompts Generation, RGB Images Generation, and RGBA Images Acquisition.

\textbf{Prompts Generation:} First, we manually prepare several instructions to guide GPT-4o \cite{achiam2023gpt} in generating appropriate text-to-image prompts. Fig. \ref{figs4} shows two types of instructions, with the content inside `\{\}' being replaceable. By continuously replacing the content within `\{\}', we eventually obtain 8,072 text-to-image prompts. Some specific generated prompts are shown in Fig. \ref{figs5}.

\textbf{RGB Images Generation:} Next, for each prompt, we use Stable Diffusion v3 \cite{esser2024scaling} to generate approximately 50 different RGB images.

\textbf{RGBA Image Acquisition:} Finally, we utilize SAM \cite{kirillov2023segment} to extract the main foreground objects from the RGB images to obtain RGBA images. Since each step of the synthesis process may produce samples that do not meet expectations, we manually filter out the unwanted samples from the RGBA images. As a result, we obtain 80,945 high-quality RGBA images. Fig.~\ref{figs6} shows some samples, and the entire dataset will be released soon.

\section{More Discussion on Background Colors}
\label{S7}
\begin{figure*}[t]
    \centering
    \includegraphics[width=\textwidth]{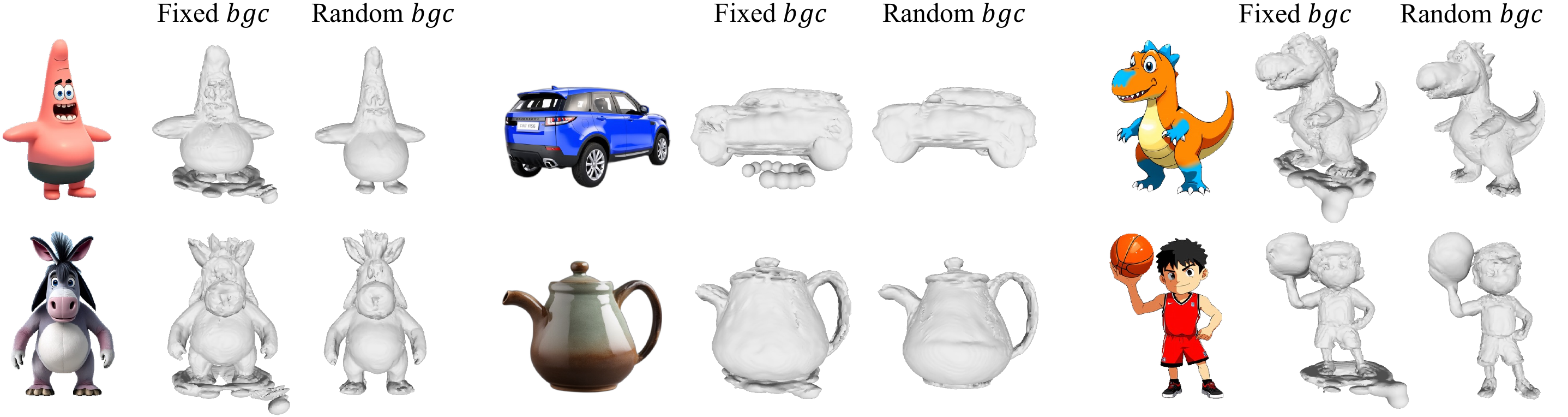}
    \caption{Ablation study of the background color.}
    \label{figs8}
\end{figure*}

To avoid PEPD frequently generating 3D Gaussians that match the background color ($bgc$), we set the background color to random values. Here, we supplement with an ablation experiment on the background color, where we convert the 3D Gaussians generated by PEPD under fixed or random background colors into meshes \cite{tang2023dreamgaussian} to observe the shape of the generated objects, as shown in Fig.~\ref{figs8}. It can be seen that setting a random background color in DD3G is crucial for generating reasonably shaped 3D objects.

\bibliography{main}

\vfill